%% file: main.tex
\documentclass[onefignum,onetabnum]{siamart220329}

\usepackage[letterpaper,centering,margin=1.45in]{geometry}

\graphicspath{{figures/}{./}}
\input{./ex_shared}

\ifpdf
\hypersetup{
  pdftitle={Efficient Neural Network Approaches for Conditional Optimal Transport with Applications in Bayesian Inference},
  pdfauthor={Zheyu Oliver Wang, Ricardo Baptista, Youssef Marzouk, Lars Ruthotto, Deepanshu Verma}
}
\fi
\begin{document}
\maketitle

% REQUIRED
\begin{abstract}
We present two neural network approaches that approximate the solutions of static and dynamic \emph{conditional optimal transport} (COT) problems. Both approaches enable conditional sampling and conditional density estimation, which are core tasks in Bayesian inference---particularly in the simulation-based (``likelihood-free'') setting.
Our methods represent the target conditional distribution as a transformation of a tractable reference distribution. Obtaining such a transformation, chosen here to be an approximation of the COT map, is computationally challenging even in moderate dimensions.
To improve scalability, our numerical algorithms use neural networks to parameterize candidate maps and further exploit the structure of the COT problem. Our static approach approximates the map as the gradient of a partially input-convex neural network. It uses a novel numerical implementation to increase computational efficiency compared to state-of-the-art alternatives.
Our dynamic approach approximates the conditional optimal transport via the flow map of a regularized neural ODE; compared to the static approach, it is slower to train but offers more modeling choices and can lead to faster sampling.
We demonstrate both algorithms numerically, comparing them with competing state-of-the-art approaches, using benchmark datasets and simulation-based Bayesian inverse problems. 
\end{abstract}

% REQUIRED

% REQUIRED

\section{Introduction} \label{sec:introduction}

Conditional sampling is a core computational challenge in many mathematical and statistical problems. A widespread motivation is Bayesian inference: given a statistical model $\pi(\bfy \vert \bfx)$ for some observations $\bfy$, parameterized by $\bfx$, one endows the parameters $\bfx$ with a prior distribution $\pi(\bfx)$ and therefore specifies the joint distribution $\pi(\bfx, \bfy) = \pi(\bfy \vert \bfx) \pi(\bfx)$. The computational goal of Bayesian inference is then to characterize the posterior distribution $\pi(\bfx \vert \bfy)$ for some realization of $\bfy$. While in some simple (conjugate) models, the posterior may lie in a tractable family of distributions, in more general settings one seeks to draw samples from the posterior, such that the expectation of any test function (yielding, for example, posterior moments or predictions) can be estimated via Monte Carlo. 

Most methods for sampling the posterior, such as Markov chain Monte Carlo (MCMC) \cite{brooks_2011handbook}, or for otherwise approximating it in some variational family (e.g., \cite{Blei_2017}), require evaluations of the unnormalized posterior density. This requirement limits such methods to the ``conventional'' setting of a tractable likelihood function and prior density---i.e., a likelihood and prior that can be evaluated pointwise. This setting includes likelihoods common in inverse problems, given by a forward operator that maps $\bfx$ to a prediction of the observations, which is then perturbed by noise. Even in this setting, producing thousands of approximately i.i.d.~samples from the posterior distribution may require millions of unnormalized posterior density evaluations.
This can be especially expensive for science and engineering applications where the forward operator entering the likelihood involves solving differential equations. 

Conditional sampling is even more challenging when the likelihood---and hence the unnormalized posterior density---is intractable to compute. One common reason is that the forward model is intrinsically stochastic or (equivalently) that it contains a set of (possibly high-dimensional) random latent parameters $\bfc$ that are not the target of inference; in this setting, computing the likelihood $\bfx \mapsto \pi(\bfy \vert \bfx)$ requires an expensive step of marginalization. In other applications, often the case in data assimilation \cite{Spantini2019CouplingTF} or inverse problems with data-driven priors \cite{Holden_2022}, the prior distribution may be known only through samples, without a closed-form density. In both of these cases, one can \emph{simulate} sample pairs $(\bfx_i, \bfy_i) \sim \pi(\bfx, \bfy)$ but not evaluate their joint density.

As a possible remedy, \textit{likelihood-free} (also called \textit{simulation-based}) inference methods~\cite{cranmer2020frontier} have seen significant recent interest.
Instead of relying on evaluations of the unnormalized posterior density, most of these methods employ some form of sample-driven \emph{measure transport}, either as a way of estimating conditional densities \cite{papamakarios_2019neural} or directly sampling from the posterior \cite{Durkan_2018}. The latter can be understood as using transport for conditional generative modeling: we seek transport maps, amortized over multiple realizations of the data $\bfy$, that push forward a given reference distribution to the target posterior distribution $\pi(\bfx|\bfy)$. Once such a map is obtained, one can readily generate i.i.d.\ samples from the target by evaluating the map on samples drawn from the reference given any realization of data $\bfy$. 
This distinguishes transport maps from methods such as MCMC, which are intrinsically serial (thus challenging to parallelize) and 
% costly to ``amortize'' over multiple realizations of the data  $\bfy$; 
typically require re-initialization for any new value of $\bfy$.

Under mild assumptions on the reference and posterior distributions, there exist infinitely many transport maps that couple the two. 
Accordingly, many parameterizations and construction methods for such transport maps have been pursued in recent literature. 
In particular, neural networks are widely used to represent transport maps, due to their expressiveness and scalability to high dimensions \cite{baptista2023conditional, Oberman_2020, dinh2017density, onken2021ot, huang2021convex}.
Yet in many settings, the target map, e.g., the map obtained in the limit of increasing expressiveness of the network, is not clearly or uniquely specified. The lack of a specific target transport leaves open a wide variety of ad hoc design choices, where these choices implicitly determine the class of transport maps being considered. Moreover, it is challenging to understand the approximation properties of resulting transports and their implications for computational cost. A more systematic way of using neural networks to represent conditional transports with a well-specified target map is desirable. 
To this end, we will seek neural network approximations of conditional \emph{optimal} transport (COT) maps, in particular, maps that are optimal under the $L^2$ transport cost. These maps are unique, under mild assumptions on the reference and the target conditionals, and are also known as conditional Brenier maps \cite{Carlier2014VectorQR}. 

This paper contributes two neural network approaches for learning the static and dynamic COT maps, respectively, through maximum likelihood training.
We select conditional Brenier maps as our target transport class and enforce their gradient-of-convex-potential structure to regularize the transport map learning problem, eliminating exploration of the unnecessary off-target choices. 
This leads directly to our first approach, the \textit{partially convex potential map} (PCP-Map), where we parameterize the map as the (partial) gradient of a partially input convex neural network (PICNN)~\cite{pmlr-v70-amos17b}. PCP-Maps, by construction, approximate static conditional Brenier maps in the limit of network expressiveness.
For our second approach, called \textit{conditional optimal transport flow} (COT-Flow), we extend our previous approach, OT-Flow~\cite{onken2021ot}, to include amortization over multiple realizations of the data. Like OT-Flow, our method considers dynamic measure transport, which has recently gained much popularity. The dynamic approach realizes transport by integrating a velocity field for a finite time. COT-Flow seeks a conditional optimal transport defined by the flow map of an ODE with constant-velocity (in a Lagrangian sense) trajectories. 
Rather than \emph{strongly} enforcing the structure of optimal transport on this velocity field, we encourage optimality of the conditional transport via appropriate penalties on the velocity. In particular, we represent the velocity as the gradient of a scalar potential parameterized by a neural network (hence obtaining a neural ODE) and penalize the squared magnitude of the velocity and violations of a Hamilton--Jacobi--Bellman PDE.

We compare the performance of our approaches with each other and against other state-of-the-art neural and non-neural approaches for conditional sampling and density estimation in terms of the computational cost of training and sampling, the quality of the resulting samples, and the complexity of hyperparameter tuning. 
We conduct numerical experiments using six UCI tabular datasets~\cite{Kelly_uci}, inverse problems involving the stochastic Lotka--Volterra model~\cite{NIPS2016_6aca9700}, and the one-dimensional shallow water equations~\cite{GATSBI}. 
Across all experiments, we observe that both PCP-Map and COT-Flow achieve competitive accuracy as measured by negative log-likelihoods, sample visualizations, and posterior calibration analyses, compared to consistent approximate Bayesian computation methods \cite{Toni_2009} and other transport or deep learning approaches~\cite{baptista2022representation, GATSBI}. 
While PCP-Map and COT-Flow enjoy comparable accuracy, the strongly-constrained PCP-Map can be trained more quickly than COT-Flow in our experiments. Also, PCP-Map relies on fewer and less influential hyperparameters than the weakly constrained COT-Flow and thus requires less hyperparameter tuning. COT-Flow, however, offers faster sampling after training due to its nearly straight-line velocity field.
We also conduct a comparative study of PCP-Map and a conditional version of the optimal transport approach proposed in~\cite{huang2021convex} and observe that our numerical implementation improves numerical accuracy and efficiency.

% Organization
The remainder of the paper is organized as follows.
\Cref{sec:background} formulates the conditional sampling problem and reviews related learning approaches.
\Cref{sec:PCP-Map} presents our partially convex potential map (PCP-Map). 
\Cref{sec:OTregularizedML} presents our conditional optimal transport flow (COT-Flow).
\Cref{sec:implementation} describes our effort to achieve reproducible results and systematic procedures for identifying effective hyperparameters for neural network training.
\Cref{sec:experiments} contains a detailed numerical evaluation of both approaches using six open-source data sets and experiments motivated by Bayesian inference for the stochastic Lotka--Volterra equation and the shallow water equations.
\Cref{sec:discussion} features a detailed discussion of our results and highlights the advantages and limitations of the presented approaches.

\section{Background and related work}\label{sec:background}
Given i.i.d.~sample pairs $\{(\bfx_i,\bfy_i)\}_{i=1}^N$ drawn from the joint distribution $\pi(\bfx,\bfy)$, with $\bfx_i \in \R^n$ and $\bfy_i \in \R^m$, our goal is to sample from $\pi(\bfx|\bfy)$ for any value of the conditioning variable $\bfy$.
In the context of Bayesian inference, $\pi(\bfx | \bfy)$ is the posterior distribution. Even when the likelihood is not available, the joint sample pairs can be obtained by first drawing $\bfx_i$ from the prior distribution $\pi(\bfx)$ and then drawing $\bfy_i$ from $\pi(\bfy | \bfx_i)$.

Under the transport-based framework, solving this conditional sampling problem amounts to learning a block-triangular map, as described in \cite{baptista2023conditional},
\begin{align}\label{eq: blocktriangular}
    T: \R^{n+m} \rightarrow \R^{n+m}, \quad
    T(\bfz_x, \bfz_y) = 
  \begin{bmatrix*}[l]
      \mathsf{h}(\bfz_y) \\
      \mathsf{g} \left ( \bfz_x, \mathsf{h}(\bfz_y) \right )
  \end{bmatrix*}
\end{align}
that pushes forward a tensor-product reference distribution $\rho_\bfz(\bfz) := \rho_{\bfz_x}(\bfz_x)\otimes\rho_{\bfz_y}(\bfz_y)$ to the target joint distribution $\pi(\bfx, \bfy) = \pi(\bfy)\pi(\bfx|\bfy)$ where $\rho_{\bfz_x}(\bfz_x)$ and $\rho_{\bfz_y}(\bfz_y)$ are defined on $\R^n$ and $\R^m$, respectively. 
More specifically, to evaluate $T(\bfz_x, \bfz_y)$, we first evaluate the mapping $\mathsf{h}$ on $\bfz_y$ and then substitute $\mathsf{h}(\bfz_y)$ into $\mathsf{g}$ to evaluate $\mathsf{g}(\bfz_x, h(\bfz_y))$. Notice that $\mathsf{h}: \R^m \to \R^m$ and $\mathsf{g}: \R^n \times \R^m \to \R^n$. If $\bfz_y\sim\rho_{\bfz_y}(\bfz_y)$ and $\bfz_x\sim\rho_{\bfz_x}(\bfz_x)$, Theorem 2.4 in~\cite{baptista2023conditional} shows that we will have $\mathsf{g}(\bfz_x, \mathsf{h}(\bfz_y))\sim\pi(\bfx|\bfy)$.
% \todo{Added reference to theorem to justify that this block-triangular map samples conditionals}. 

We can simplify this construction by choosing $\mathsf{h}$ to be the identity function and setting the reference distribution to be $\pi(\bfy)\otimes\rho_{\bfz_x}(\bfz_x)$
% \todo{This may look nicer with the arguments for all densities}.\owtd{done!} 
The reference marginal $\rho_{\bfz_x}(\bfz_x)$ can be chosen as any tractable distribution, e.g., a standard Gaussian. We can use the conditioning variable $\bfy$ as the second argument of $\mathsf{g}$. The block triangular map becomes
\begin{align}
    T(\bfz_x, \bfy) = 
    \begin{bmatrix*}[l]
      \bfy \\
      \mathsf{g} (\bfz_x, \bfy)
    \end{bmatrix*}.
\end{align}
With this construction, we can sample from the conditional distribution $\pi(\bfx|\bfy)$ simply by evaluating the mapping $g: \bfz_x \mapsto \mathsf{g}(\bfz_x, \bfy)$ for $\bfz_x \sim \rho_{\bfz_x}(\bfz_x)$. We will focus on such a $g$ in this paper, and call it the \textit{conditional generator}. We view $g$ as a mapping from $\R^n \to \R^n$, parameterized by $\bfy$, and will write $g(\bfz_x; \bfy)$, for $\bfz_x \in \R^n$, to make this interpretation clear.
We will also use $g^{-1}(\cdot \, ; \bfy)$ to denote the inverse of this mapping (when it exists) for any given value of $\bfy$.  In other words, $g \left ( g^{-1} (\bfx ; \bfy) ; \bfy \right ) = \bfx$.

There are numerous ways of parameterizing, constraining, and learning the conditional generator. For example, \cite{baptista2022representation, Jaini19} use transformations of polynomials to approximate $g$. While linear spaces (such as polynomial spaces) are amenable to analysis and have well-understood approximation properties, they typically suffer from the computational difficulties of basis selection and the curse of dimensionality.
In the deep generative modeling community \cite{RuthottoHaber2021}, neural networks are instead used to parameterize $g$.
This paper will focus on deep generative modeling approaches that learn {invertible} and differentiable generators; as we will explain shortly, these properties enable exact density evaluations and, hence, maximum likelihood training. This choice thus excludes other conditional generative models popular in machine learning, such as conditional VAEs~\cite{NIPS2015_8d55a249} and conditional GANs~\cite{Mirza2014ConditionalGA, Liu2021WassersteinGL}, as they do not employ invertible maps or allow density evaluations. In such methods, when the latent code $\bfz$ has a dimension less than $n$, the pushforward distribution, in fact, does not have a density with respect to the Lebesgue measure on $\R^n$. Therefore, these methods must be trained by methods other than maximum likelihood.

Given a conditional generator $g(\cdot \, ; \bfy)$ that is a diffeomorphism for any value of $\bfy$, we can write the pushforward of the reference density $\rho_{\bfz_x}$ by $g(\cdot \, ; \bfy)$ via the change-of-variables formula:
\begin{equation} 
\label{eq:changeOfVariables}
\begin{split}
g(\cdot ; \bfy)_\sharp \rho_{\bfz_x}(\bfx) & \coloneqq \rho_{\bfz_x}\left(g^{-1}(\bfx ; \bfy)\right) \left|\det\nabla_\bfx g^{-1}(\bfx ;\bfy)\right|.
\end{split}
\end{equation} 
Then we can use $g(\cdot ; \bfy)_\sharp \rho_{\bfz_x}(\bfx)$ to approximate the conditional density $\pi(\bfx \vert \bfy)$, i.e., to seek
\begin{equation}
g(\cdot ; \bfy)_\sharp \rho_{\bfz_x}(\bfx) \approx \pi(\bfx \vert \bfy).
\label{eq:condmatching}
\end{equation}
The explicit density in \eqref{eq:changeOfVariables} is quite useful in this regard; for instance, it allows us to identify a good $g$ within a given class of diffeomorphisms via maximum likelihood estimation.

In general, there are infinitely many diffeomorphisms that can achieve equality in \eqref{eq:condmatching}. 
One class of diffeomorphisms for smooth densities is given by monotone triangular maps \cite{MarzoukEtAl2016, wang_2022}, which include the canonical Knothe--Rosenblatt rearrangement; the latter achieves equality in \eqref{eq:condmatching} for any absolutely continuous target. Triangular maps allow easy inversion and Jacobian determinant evaluations, though they are sensitive to variable ordering.
A different class of candidate diffeomorphisms is given by conditional normalizing flows~\cite{Lu2020CondGLOW, MaskedAF}; these are specific neural network representations of diffeomorphisms on $\R^n$, parameterized by $\bfy$, that also allow easy inversion and evaluations of the Jacobian determinant. A vast array of architectures for such flows, in the conditional and unconditional cases, have been proposed in the literature \cite{papamakarios2021normalizing}, and there even exist universal approximation results for certain common architectures \cite{Teshima_2020}. 

Early normalizing flows \cite{dinh2017density, InverseAF, MaskedAF} directly modeled the displacement of points from the reference to the target distribution. They can be thus viewed as a ``static'' approach to measure transport. Most of these flows do not aim to approximate any particular canonical transport, although many (e.g., autoregressive flows) use restricted classes of triangular maps as a building block.
Continuous normalizing flows, as proposed in \cite{Chen_2018,grathwohl2019ffjord}, on the other hand,
represent the diffeomorphism of interest by the finite-time flow map of an ODE \cite{Marzouk_2023} whose velocity field is parameterized by a deep neural network. These flows thus can be viewed as a ``dynamic'' form of measure transport.
Generic continuous normalizing flows also lack a specific approximation target; they can, in principle, represent infinitely many transport maps that are consistent with \eqref{eq:condmatching}, with no direct control over what map is realized. To resolve this ill-posedness \cite{zhang_2023}, recent works have sought to tailor continuous NFs towards specific target transports.
In particular, there has been considerable interest in targeting \textit{optimal transport} with quadratic cost (i.e., the Brenier map) to take advantage of its properties. Several dynamic measure transport approaches drive their flows towards optimal transport by adding regularization penalties to the training objective \cite{onken2021ot, Oberman_2020, zhang2018mongeampere, YangKarniadakis2019PotentialFlow}.
Static density estimation and generative modeling approaches that seek to approximate the Brenier map include \cite{huang2021convex,bunne2022supervised,baptista2023conditional}; we will discuss \cite{huang2021convex} and \cite{bunne2022supervised} further below. 
\Cref{fig:related_work} provides a schematic overview of various (conditional and unconditional) measure transport approaches, arranged by the approximation target (if any) and whether they are static or dynamic.

\begin{figure}[t]
\centering
\begin{tikzpicture}[scale=0.88]
    \definecolor{lightred}{rgb}{1.0, 0.35, 0.37}
    \definecolor{lightblue}{rgb}{0.19, 0.45, 0.95}
    \definecolor{darkgray}{rgb}{0.35, 0.35, 0.35}
    % outline
    % KR circ
    \draw[dashed, lightred] (0.55, 0.8) ellipse (0.8 and 0.45);
    \fill[lightred] (1.43, 0.8) circle (0.14cm);
    % OT circ
    \draw[dashed, lightred] (2.83, -1.6) ellipse (1.4 and 1.1);
    \draw[dashed, lightred] (-0.41, -1.9) ellipse (1.7 and 1.1);
    \fill[lightred] (1.34, -1.8) circle (0.16cm);
    \draw
    (1.36, -1.6) node [text=lightred,above] {\textbf{\large OT}}
    (1.4, 1.05) node [text=lightred,above] {\textbf{ KR}}

    % Static
    (-2.5, 2) node [text=lightblue,above] {\large\textbf{Static}}
    (-2.4, -3.3) node [text=black,above] {\small Glow\cite{GlowGF}}
    (-1.2, -0.45) node [text=black,above] {\small Conditional Glow\cite{Lu2020CondGLOW}}
    (0, 2.3) node [text=black,above] {\small RealNVP\cite{dinh2017density}}
    (-2.8, -1.3) node [text=black,above] {\small NAF\cite{NeuralAF}}
    (-3, -2) node [text=black,above] {\small IAF\cite{InverseAF}}
    (0.1, 1.6) node [text=black,above] {\small RQ-NSF\cite{durkan2019splineflow}}
    (-3, -2.7) node [text=black,above] {\small MAF\cite{MaskedAF}} 
    (-2.31, 0.62) node [text=black, above] {\small UMNN\cite{Wehenkel2019UnconstrainedMN}}

    % Dynamic
    (5.2, -2.9) node [text=lightblue,above] {\large \textbf{Dynamic}}
    (5, -0.2) node [text=black,above] {\small FFJORD\cite{grathwohl2019ffjord}}
    (5.4, -1.8) node [text=black,above] {\small CNF \cite{Chen_2018}}
    (3.4, 1) node [text=black,above] {\small ANODE \cite{Dupont_2019}}
    
    % Dynamic OT
    (2.95, -1.5) node [text=black, above] {\small RNODE\cite{Oberman_2020}}
    (2.9, -2) node [text=black, above] {\small OT-Flow\cite{onken2021ot}}
    (2.8, -2.4) node [text=black, above] {\small \textbf{COT-Flow}}
    % Static OT
    (-0.48, -1.6) node [text=black, above] {\small CP-Flow\cite{huang2021convex}}
    (-0.48, -2) node [text=black, above] {\small \textbf{PCP-Map}}
    (-0.48, -2.45) node [text=black, above] {\small CondOT\cite{bunne2022supervised}}
    (-0.48, -2.9) node [text=black, above] {\small MGAN\cite{baptista2023conditional}}
    % KR
    (0.54, 0.46) node [text=black,above] {\small ATM\cite{baptista2022representation}}
    
    (-4,-3.3) rectangle (6.8, 3.1) (4.7, 2.3) node [text=darkgray,above] {\textbf{\large Diffeomorphic}};
    
    \draw[solid, color=lightblue] (1.3,-3.3)--(1.5, 3.1);
\end{tikzpicture}
\caption{Schematic overview of related measure transport approaches, all of which produce diffeomorphic transport maps. Approaches are first separated into static versus dynamic, then grouped by their approximation target assumptions.
The red dots represent specific, canonical transport maps: ($L^2$) optimal transport (OT) and the Knothe--Rosenblatt (KR) transport. The surrounding ellipses capture methods that seek to approximate these canonical maps.}
\label{fig:related_work}
\end{figure}

As discussed in~\Cref{sec:introduction}, this paper focuses on neural network approaches to learning COT maps. A close relative to our PCP-Map construction is the CP-Flow approach of~\cite{huang2021convex}. CP-Flow is primarily designed to approximate solutions of the joint (rather than conditional) optimal transport problem. A variant of CP-Flow defines a variational autoencoder (VAE) suitable for variational Bayesian inference. But extensions of CP-Flow to COT are also natural; though not described in the paper, a script in the GitHub repository associated with~\cite{huang2021convex} enables CP-Flow to approximate the COT problem for a simple Gaussian mixture target distribution in a similar fashion to PCP-Map. We, therefore, compare PCP-Map to conditional CP-Flow numerically; see \Cref{sub:comp_cp_flow}. In these experiments, we find that CP-Flow either fails to provide a solution or is considerably slower than PCP-Map. We discuss possible reasons in \Cref{sub:comp_cp_flow}, but note here some distinctions between (conditional) CP-Flow and PCP-Map: the latter uses a simplified transport map architecture, new automatic differentiation tools that avoid the need for stochastic log-determinant estimators, and a projected gradient method to enforce non-negativity constraints on parts of the weights. 

Another close relative of PCP-Map is the CondOT approach in~\cite{bunne2022supervised}. The main distinction between CondOT and PCP-Map is in the definition of the learning problem. 
The former solves the $W_2$ COT problem in an adversarial manner, which leads to a challenging stochastic saddle point problem. Our approach instead uses a maximum likelihood estimation, which yields a simpler, direct minimization problem. 

The COT-Flow developed in this paper approach extends the dynamic OT formulation in~\cite{onken2021ot} to enable conditional sampling and density estimation. 
The resulting approach is similar to the recent work~\cite{Pandey2024}, which uses an RKHS approach to approximate the velocity and bases the training objective function on maximum mean discrepancy (MMD) instead of maximum likelihood.

\section{Partially convex potential maps (PCP-Map)}

\label{sec:PCP-Map}
Our first approach, PCP-Map, approximates the conditional optimal transport map as the \emph{partial} gradient of a \emph{{partially} input convex neural network (PICNN)}~\cite{pmlr-v70-amos17b}.
This design enforces the structure of conditional Brenier maps.
As common in normalizing flows, we learn the parameters of the neural network by maximizing the likelihood of the training samples.

\paragraph{Training problem}
Given samples from the joint distribution, we train the conditional generator $\gx$ by solving the maximum likelihood problem
\begin{equation}\label{eq:TrainProb}
	\min_{\gx^{-1} \in \mathcal{G}}\; J_{{\rm NLL}}[\gx^{-1}],
\end{equation}
where the objective $J_{{\rm NLL}}$ is
the expected negative log-likelihood functional
\begin{equation}\label{eq:JNNLx}
	J_{{\rm NLL}}[\gx^{-1}] = 
	\mathbb{E}_{\pi(\bfx,\bfy)} \left[ \hf \left\| \gx^{-1}(\bfx; \bfy) \right\|^2 - \log \det \nabla_\bfx \gx^{-1}(\bfx;\bfy)  \right].
\end{equation}
This objective arises from choosing $\rho_{\bfz_x}$ to be a standard Gaussian; maximizing it seeks to make $\gx^{-1}(\bfx; \bfy)$ also standard Gaussian.
Since the learning problem in~\eqref{eq:JNNLx} only involves the \emph{inverse conditional generator} $\gx^{-1}$, in \eqref{eq:TrainProb} we follow~\cite{baptista2023conditional} and directly seek conditional generators whose inverses have the structure of conditional Brenier maps, i.e.,
\begin{equation} \label{eq:breniertypeMaps}
    \mathcal{G} = \left\{\bfx \mapsto \nabla_{\bfx} G(\bfx, \bfy) \ \text{s.t.} \  G:\R^n\times\R^m\rightarrow\R \text{ is convex in $\bfx$} \right\}.
\end{equation}
These parameterizations are equivalent: if, for any fixed $\bfy$, $g(\cdot; \bfy)$ is the gradient of a convex potential, then $g(\cdot; \bfy)^{-1}$ is also the gradient of a (different) convex potential by Brenier's theorem, and vice versa.
While this choice avoids inverting the generator during training, sampling requires inverting the learned map. Note that the inverse conditional generator pushes forward the conditional distribution $\pi(\bfx|\bfy)$ to the reference distribution. We note that the negative log-likelihood $J_{\rm NLL}$ also agrees (up to an additive constant) with the Kullback--Leibler (KL) divergence from the pushforward $g(\cdot ; \bfy)_\sharp \rho_{\bfz_x}$ to the conditional distribution $\pi(\bfx|\bfy)$, in expectation over the conditioning variable $\bfy \sim \pi(\bfy)$.

Theoretically speaking, Brenier's theorem ensures the existence of a unique monotone transport map among all maps written as the gradient of a convex potential; see~\cite{Brenier} for the original result and \cite{Carlier2014VectorQR} for conditional transport maps.
Theorem 2.3 in~\cite{Carlier2014VectorQR} also shows that $\gx^{-1}$ is \textit{optimal} in the sense that among all maps that match the distributions, it minimizes the integrated $L_2$ transport costs
\begin{equation}\label{eq:P_OT}
	P_{\rm OT}[\gx] = \Ex_{\pi(\bfx,\bfy)} \left[  \| \bfx - \gx^{-1}(\bfx;\bfy)\|^2
	\right].
\end{equation} 

\paragraph{Neural Network Representation}
We directly parameterize $\gx^{-1}_\theta$ as the partial gradient of a scalar-valued PICNN-based potential $\Ginv_{\theta}: \R^n\times\R^m \to \R$ that depends on weights $\theta$. 
A PICNN is a feed-forward neural network that is specifically designed to ensure convexity in part of its inputs; see the original work that introduced this neural network architecture in~\cite{pmlr-v70-amos17b} and its use for generative modeling in~\cite{huang2021convex}.
To the best of our knowledge, investigating whether PICNNs are universal approximators of partially input convex functions remains an open issue (also see \cite[Section~IV]{Kim-Kim-2022}) that is beyond the scope of our paper; a perhaps related result for fully input convex neural networks is given in~\cite[Appendix C]{huang2021convex}.

To ensure the strict monotonicity of $\gx^{-1}_\theta$, we construct $\Ginv_\theta$ to be strictly convex as a linear combination of a scalar-valued PICNN and a positive definite quadratic term. That is,
\begin{equation}\label{eq: strictcvxpot}
	\Ginv_{\theta}(\bfx,\bfy) = \sigma_{\rm softplus}(\gamma_1) \cdot w_K + (\ReLU(\gamma_2) + \sigma_{\rm softplus}(\gamma_3)) \cdot \frac{1}{2} \|\bfx\|^2,
\end{equation}
where $\gamma_1, \gamma_2,$ and $\gamma_3$ are scalar parameters that are re-parameterized via the soft-plus function $\sigma_{\rm softplus}(x) = \log(1+\exp(x))$ and the ReLU function $\ReLU(x) = \max\{0,x\}$ to ensure strict convexity of $\Ginv_\theta$. Here, $w_K$ is the output of a $K$-layer scalar-valued PICNN and is computed through forward propagation through the layers $k=0,\ldots,K-1$ starting with the inputs $\bfv_0 = \bfy$ and $\bfw_0=\bfx$
\begin{equation}\label{eq:PICNN}
	\begin{split}
		& \bfv_{k+1} = \sigma^{(v)} \; \left( \bfL^{(v)}_k\bfv_k \; + \; \bfb^{(v)}_k \right),\\
		& \bfw_{k+1} = \sigma^{(w)} \; \left( \ReLU\left(\bfL^{(w)}_k\right)\left(\bfw_k \odot \; \ReLU\left(\bfL^{(wv)}_k\bfv_k +  \bfb^{(wv)}_k\right)\right) + \right. \\
		&\left. \quad\quad \bfL^{(x)}_k \left(\bfx \; \odot \; \left(\bfL^{(xv)}_k\bfv_k + \bfb^{(xv)}_k\right)\right) + \bfL^{(vw)}_k\bfv_k + \bfb^{(w)}_k \right).  \\
	\end{split}
\end{equation}
Here, $\bfv_k$, termed contexts, are layer activations of input $\bfy$, and  $\odot$ denotes the element-wise Hadamard product. 
We implement the non-negativity constraints on $\bfL^{(w)}_k$ and $\bfL^{(wv)}_k\bfv_k \; + \; \bfb^{(wv)}_k$ via the ReLU activation and set  $\sigma^{(w)}$ and $\sigma^{(v)}$ to the softplus and ELU functions, respectively, where
\begin{align*}
    \sigma_{\rm ELU}(x) = 
    \begin{cases}
        x & \text{if}\;x>0\\
        e^x - 1 &\text{if}\;x\leq 0
    \end{cases}.
\end{align*}

We list the dimensions of the weight matrices in~\Cref{tab:PICNNarchitecture}. The sizes of the bias terms equal the number of rows of their corresponding weight matrices. The trainable parameters are
\begin{equation}
\begin{split}
	\theta = (\gamma_{1:3}, &\bfL^{(v)}_{0:K-2}, \bfb^{(v)}_{0:K-2}, \bfL^{(vw)}_{0:K-1}, \bfL^{(w)}_{0:K-1}, \bfb^{(w)}_{0:K-1}, \\
 & \bfL^{(wv)}_{0:K-1}, \bfb^{(wv)}_{0:K-1}, \bfL^{(xv)}_{1:K-1}, \bfb^{(xv)}_{1:K-1}, \bfL^{(x)}_{1:K-1}).   
\end{split}
\end{equation}
Using properties for the composition of convex functions~\cite{DCP}, it can be verified that the forward propagation in~\eqref{eq:PICNN} defines a function that is convex in $\bfx$, but not necessarily in $\bfy$ (which is not needed), as long as $\sigma^{(w)}$ is convex and non-decreasing. 
\begin{table}[t]
	\centering
    \small
	\caption{PCP-Map, \Cref{sec:PCP-Map}. Parameter dimensions of a $K$-layer PICNN architecture from~\eqref{eq:PICNN}. Since the dimensions of the inputs, $\bfx\in\R^n$ and $\bfy \in\R^m$, are given, and the network outputs a scalar, we vary only the depth, $K$, feature width, $w$, and context width, $u$, in our experiments.} \label{tab:PICNNarchitecture}
	\begin{tabular}{|c|cccccc|}
        \hline
		$k$, layer & size($\bfL_k^{(v)}$)  & size($\bfL_k^{(vw)}$)  & size($\bfL_k^{(w)}$)  & size($\bfL_k^{(wv)}$)  & size($\bfL_k^{(xv)}$)  & size($\bfL_k^{(x)}$) \\[2pt]\hline\hline
		$0$ & ($u$, $m$) & ($w$, $m$) & ($w$, $n$) & ($n$, $m$) & 0 & 0\\
		$1$ & ($u$, $u$) & ($w$, $u$) & ($w$, $w$) & ($w$, $u$) & ($n$, $u$) & ($w$, $n$)\\
		$\vdots$ & $\vdots$ & $\vdots$ & $\vdots$ & $\vdots$ & $\vdots$ & $\vdots$\\
		$K-2$ & ($u$, $u$) & ($w$, $u$) & ($w$, $w$) & ($w$, $u$) & ($n$, $u$) & ($w$, $n$)\\
		$K-1$ & 0 & (1, $u$) & (1, $w$) & ($w$, $u$)& ($n$, $u$) & ($w$, $n$)\\
        \hline
	\end{tabular}
\end{table}

To compute the log determinant of $\gx^{-1}_\theta$, we use vectorized automatic differentiation to obtain the Hessian of $\Ginv_{\theta}$ with respect to its first input and then compute its eigenvalues.
This is feasible when the Hessian's dimension is moderate; e.g., in our experiments, it is less than one hundred. 
We use efficient implementations of these methods that parallelize the computations over all the samples in a batch.

Our algorithm enforces the non-negativity constraint by projecting the parameters into the non-negative orthant after each optimization step using ReLU. Thereby, we alleviate the need for re-parameterization, for example, using the softplus function in~\cite{pmlr-v70-amos17b}.
Another novelty introduced in PICNN is that we utilize trainable affine layer parameters $\bfL^{(v)}_k$ and a context width $u$ as a hyperparameter to increase the expressiveness over the conditioning variables, which are pivotal to characterizing conditional distributions; existing works such as~\cite{huang2021convex}  set $\bfL^{(v)}_{1:K-2} = \bfI$. 

\paragraph{Sample generation}
Due to the neural network parameterization, there is generally no closed-form relation between $\gx_\theta$ and $\gx^{-1}_\theta$. 
As in \cite{huang2021convex}, we approximate the inverse of $\gx^{-1}_\theta$ during sampling as the Legendre-Fenchel dual. That is, given a reference sample $\bfz_x \sim \rho_{\bfz_x}(\bfz_x)$ and a conditioning variable $\bfy\sim\pi(\bfy)$, we generate samples by solve the convex optimization problem
\begin{equation}\label{eq: pcp_sample_optimprob}
	\bfv^* = \argmin_{\bfv} \;\Ginv_{\theta}(\bfv, \bfy) - \bfz_x^\top\bfv.
 \end{equation}
Due to the strict convexity of $\Ginv_{\theta}$ in its first argument, the first-order optimality conditions gives 
\begin{align*}
    \bfv^* \approx \nabla_{\bfz_x} \Ginv^{-1}_{\theta}(\bfz_x; \bfy) = \gx_\theta(\bfz_x; \bfy) \sim \pi(\bfx|\bfy).
\end{align*}
Numerically, the optimization problem (\ref{eq: pcp_sample_optimprob}) is solved approximately using $l$-BFGS with line search under strong Wolfe conditions. The optimization is terminated when a pre-set tolerance is reached.

\paragraph{Hyperparameters}
In our numerical experiments, we vary only three hyperparameters to adjust the complexity of the architecture. As described in~\Cref{sec:implementation} we select the depth $K$, feature width $w$, and context width $u$ from the values in~\Cref{tab:hyparam}. 

\section{Conditional OT-Flow (COT-Flow)} \label{sec:OTregularizedML}

In this section, we approximate the COT map $g$ with a continuous normalizing flow (CNF) ~\cite{Chen2019ResidualFF,grathwohl2019ffjord} whose time-dependent \textit{velocity field} is the \emph{partial} gradient of a scalar-valued neural network. Similarly to the previous section, we learn the parameters of the network through maximum likelihood training but add an OT regularization term to enforce uniqueness and favor simple generators with straight trajectories.  To this end, we extend the OT-regularized CNF in~\cite{onken2021ot} to the conditional setting.

\paragraph{Training Problem}
Following the general approach of CNFs, we represent a $g$ that pushes forward the reference distribution $\rho_{\bfz_x}(\bfz_x)$ to the target distribution $\pi(\bfx|\bfy)$ via the flow map of an ODE \cite{Marzouk_2023}.
That is, we define $\gx(\bfz_x;\bfy)=\bfu(1)$ as the terminal state of the ODE
\begin{equation}\label{eq:CNF}
	\frac{d}{dt} \bfu(t) = v(t,\bfu(t); \bfy), \quad t \in (0,1], \quad \bfu(0)=\bfz_x,
\end{equation}
where the evolution of $\bfu\colon [0,1] \to \R^n$ depends on the velocity field $v\colon [0,1]\times \R^n \times \R^m \to \R^n$ and will be parameterized with a neural network below.
When the velocity is trained to minimize the negative log-likelihood loss in~\eqref{eq:JNNLx}, the resulting mapping is called a continuous normalizing flow~\cite{grathwohl2019ffjord}.
We add $\bfy$ as an additional input to the velocity field to enable conditional sampling. Consequently, the resulting flow map and generator $\gx$ depend on $\bfy$.

One advantage of defining the generator through an ODE is that the loss function can be evaluated efficiently for a wide range of velocity functions. 
Recall that the loss function requires the inverse of the generator and the log-determinant of its Jacobian.
For sufficiently regular velocity fields, the inverse of the generator can be obtained by integrating backward in time. 
To be precise, we define $\gx^{-1}(\bfx;\bfy) = \bfp(0)$ where $\bfp : [0,1] \to \R^n$ satisfies~\eqref{eq:CNF} with the terminal condition $\bfp(1)=\bfx$.
As derived in~\cite{zhang2018mongeampere, grathwohl2019ffjord, YangKarniadakis2019PotentialFlow}, constructing the generator through an ODE also simplifies computing the log-determinant of the Jacobian, i.e., 
\begin{equation}\label{eq:JacboiIdentity}
\log \det \nabla_\bfx \gx^{-1}(\bfx;\bfy) = \int_{0}^1 {\rm trace}\left(\nabla_\bfp v(t,\bfp(t); \bfy)\right) dt.
\end{equation}

Penalizing transport costs during training leads to theoretical and numerical advantages; see, for example, ~\cite{zhang2018mongeampere,YangKarniadakis2019PotentialFlow,onken2021ot, Oberman_2020}.
Hence, we consider the OT-regularized training problem
\begin{equation}\label{eq:OTflowProb}
	\min_v \;J_{\rm NLL}[g^{-1}] + \alpha_1 P_{\rm DOT}[v]
\end{equation}
where $\alpha_1>0$ is a regularization parameter that trades off matching the distributions (for $\alpha_1 \ll 1$) and minimizing the transport costs (for $\alpha_1\gg 0$) given by the dynamic transport cost penalty
\begin{equation}\label{eq:P_DOT}
	P_{\rm DOT}[v] = \Ex_{\pi(\bfx,\bfy)} \left[ \int_0^1 \frac{1}{2} \|v(t,\bfp(t); \bfy)\|^2 dt\right].
\end{equation}
This penalty is stronger than the static counterpart in~\eqref{eq:P_OT}, in other words
\begin{equation}
	P_{\rm OT}[g] \leq P_{\rm DOT}[g] \quad \forall g : \R^{n} \to \R^n.
\end{equation}
However, the values of both penalties agree when the velocity field in~\eqref{eq:CNF} is constant along the trajectories, which is the case for the optimal transport map. 
Hence, we expect the solution of the dynamic problem to be close to that of the static formulation when $\alpha_1$ is chosen well.

To provide additional theoretical insight and motivate our numerical method, we note that ~\eqref{eq:OTflowProb} is related to the potential mean field game (MFG)
\begin{equation}\label{eq:dynamicTrainingProblem}
	\begin{split}
		\min_{\varrho,v} \quad & \int_{\R^m}\int_{\R^n}   \left(- \log \varrho(1,\bfx; \bfy) \pi(\bfx,\bfy) + \alpha_1\int_0^1  \frac{1}{2} \|v(t,\bfx;\bfy)\|^2 \varrho(t,\bfx;\bfy) dt \right) d\bfx d\bfy \\
		\text{subject to} &\quad
		\partial_t \varrho(t,\bfx;\bfy) + \nabla_{\bfx}\cdot\left(\varrho(t,\bfx; \bfy) v(t,\bfx; \bfy)\right)=0, \quad t \in (0,1] \\ 
		& \quad \varrho(0,\bfx;\bfy) = \rho_{\bfz_x}(\bfx).
	\end{split}    
\end{equation}
Here, the terminal and running costs in the objective functional are the $L_2$ anti-derivatives of $J_{\rm NLL}$ and $P_{\rm DOT}$, respectively, and the continuity equation is used to represent the density evolution under the velocity $v$.
To be precise,~\eqref{eq:OTflowProb} can be seen as a discretization of~\eqref{eq:dynamicTrainingProblem} in the Lagrangian coordinates defined by the reversed ODE~\eqref{eq:COTflow}.
Note that both formulations differ in the way transport costs are measured: \eqref{eq:P_DOT} computes the cost of pushing the conditional distribution to the Gaussian, while~\eqref{eq:dynamicTrainingProblem} penalizes the costs of pushing the Gaussian to the conditional. 
While these two terms do not generally agree, they coincide for the $L_2$ optimal transport map; see~\cite[Corollary 2.5.13]{figalli2021invitation} for a proof for unconditional optimal transport.
For more insights into MFGs and their relation to optimal transport and generative modeling, we refer to our previous work~\cite{RuthottoEtAl2020MFG} and the more recent work~\cite[Sec 3.4]{zhang_2023}.
The fact that the solutions of the microscopic version in~\eqref{eq:OTflowProb} and the macroscopic version in~\eqref{eq:dynamicTrainingProblem} agree is remarkable and was first shown in the seminal work~\cite{LasryLions2007}.

By Pontryagin's maximum principle,  the solution of~\eqref{eq:OTflowProb} satisfies the feedback form
\begin{equation}
	v(t,\bfp(t);\bfy) = - \frac{1}{\alpha_1}\nabla_\bfp \Phi(t,\bfp(t);\bfy)
\end{equation}
where $\Phi: [0,1]\times \R^n \times \R^m \to \R$ and $\Phi(\cdot,\cdot;\bfy)$ can be seen as the value function of the MFG and, alternatively, as the Lagrange multiplier in~\eqref{eq:dynamicTrainingProblem} for a fixed $\bfy$.
Therefore, we model the velocity as a conservative vector field as also proposed in~\cite{onken2021ot,RuthottoEtAl2020MFG}.
This also simplifies the computation of the log-determinant since the Jacobian of $v$ is symmetric, and we note that
\begin{equation}
	{\rm trace}\nabla_\bfp \left(\nabla_\bfp \Phi(t,\bfp(t);\bfy) \right) = \Delta_\bfp \Phi(t,\bfp(t);\bfy).
\end{equation}

Another consequence of optimal control theory is that the value function satisfies the Hamilton Jacobi Bellman (HJB) equations
\begin{equation}\label{eq:HJBt}
	\partial_t \Phi(t,\bfx;\bfy) - \frac{1}{2\alpha_1}\|\nabla_\bfx \Phi(t,\bfx;\bfy)\|^2 = 0, \quad t \in [0,1)
\end{equation}
with the terminal condition
\begin{equation}
	\quad \Phi(1,\bfx;\bfy) = - \frac{\pi(\bfx,\bfy)}{\varrho(1,\bfx;\bfy)},
\end{equation}
which is only tractable if the joint density  $\pi(\bfx,\bfy)$ is available.
These $n$-dimensional PDEs are parameterized by $\bfy\in\R^m$.

\paragraph{Neural network approach}
We parameterize the value function, $\Phi$, with a scalar-valued neural network.
In contrast to the static approach in the previous section, the choice of function approximator is more flexible in the dynamic approach.
The approach can be effective as long as $\Phi$ is parameterized with any function approximation tool that is effective in high dimensions, allows efficient evaluation of its gradient and Laplacian, and the training problem can be solved sufficiently well.
For our numerical experiments, we use the architecture considered in~\cite{onken2021ot} and model $\Phi$ as the sum of a simple feed-forward neural network and a quadratic term. That is, % i.e., we use
\begin{equation}
	\Phi_{\theta}(\bfq) = {\rm NN}_{\theta_{\rm NN}}(\bfq) + Q_{\theta_{\rm Q}}(\bfq), \quad \text{with } \quad \bfq = (t,\bfx;\bfy), \quad \theta = (\theta_{\rm NN}, \theta_{\rm Q}).
\end{equation}
As in~\cite{onken2021ot}, we model the neural network as a two-layer residual network of width $w$ that reads
\begin{equation}
	\begin{split}
		\bfh_0 & = \sigma(\bfA_0 \bfq + \bfb_0 )\\
		\bfh_1 & = \bfh_0 + \sigma(\bfA_1, \bfh_0 + \bfb_1)\\
		{\rm NN}_{\theta_{\rm NN}}(\bfq) & = \bfa^\top \bfh_1
	\end{split}
\end{equation}
with trainable weights $\theta_{\rm NN} = (\bfa, \bfA_0, \bfb_0, \bfA_1, \bfb_1)$ where $\bfa\in\R^w, \bfA_0 \in \R^{(w \times (m+n+1)}$, $\bfb_0 \in \R^w, \bfA_1 \in \R^{w\times w}$, and $\bfb_1 \in \R^w$. 
The quadratic term depends on the weights $\theta_{\rm Q} =(\bfA, \bfb, c)$ where $\bfA\in\R^{(n+m+1)\times r}, \bfb \in \R^{m+n+1}, c \in \R$ and is defined as
\begin{equation}
	Q_{\theta_{\rm Q}}(\bfq) = \hf \bfq^\top (\bfA \bfA^\top) \bfq + \bfb^\top \bfq + c.
\end{equation}
Adding the quadratic term provides a simple and efficient way to model affine shifts between the distributions.
In our experiments $r$ is chosen to be $\min(10, n+m+1)$.

\paragraph{Training problem}
In summary, we obtain the training problem 
\begin{equation}\label{eq:COTflow}
	\begin{split}
		\min_\theta & \quad  J_{\rm NLL}[g_\theta^{-1}] + \alpha_1 P_{\rm DOT}[\nabla_\bfp \Phi_\theta] + \alpha_2 P_{\rm HJB}[\Phi_\theta]\\
		\text{ where }&\quad  g_\theta^{-1}(\bfx;\bfy) = \bfp(0)\; \text{and} \; \frac{d}{dt} \bfp(t) = -\frac{1}{\alpha_1} \nabla_\bfp \Phi_\theta(t,\bfp(t); \bfy),\; t\in(0,1], \; \bfp(1)=\bfx.
	\end{split}
\end{equation}
Here, $\alpha_2 \geq 0$ controls the influence of the HJB penalty term from~\cite{onken2021ot}, which reads
\begin{equation}
	P_{\rm HJB}[\Phi] = \Ex_{\pi(\bfx,\bfy)} \left[ \int_0^1 \left| \partial_t \Phi(t,\bfp(t);\bfy) - \frac{1}{2\alpha_1}\|\nabla_\bfp \Phi(t,\bfp(t);\bfy)\|^2 \right| dt\right]. 
\end{equation}
When $\alpha_1$ is chosen so that the minimizer of the above problem matches the densities exactly, the solution is the optimal transport map. 
In this situation, the relationship between the value function, $\Phi_\theta$, and the optimal potential, $\Ginv_\theta$, is given by
\begin{equation}
    \Ginv_\theta(\bfx,\bfy) + C = \hf \bfx^\top \bfx + \frac{1}{\alpha_1} \Phi_\theta(1,\bfx;\bfy).
\end{equation}
for some constant $C\in\R$.
To train the COT-Flow, we use a discretize-then-optimize paradigm.
In our experiments, we use $n_t$ equidistant steps of the Runge-Kutta-4 scheme to discretize the ODE constraint in~\eqref{eq:COTflow} and apply the Adam optimizer to the resulting unconstrained optimization problem.
Note that following the implementation by~\cite{onken2021ot}, we enforce a box constraint of $[-1.5, 1.5]$ to the network parameters $\theta_{NN}$.
Since the velocities defining the optimal transport map will be constant along the trajectories, we expect the accuracy of the discretization to improve as we get closer to the solution.

\paragraph{Sample generation}
After training, we draw i.i.d.\thinspace samples from the approximate conditional distribution for a given $\bfy$ by sampling $\bfz_x \sim \rho_{\bfz_x}(\bfz_x)$ and solving the ODE~\eqref{eq:CNF}.
Since we train the inverse generator using a fixed number of integration steps, $n_t$, it is interesting to investigate how close our solution is to the continuous problem. One indicator is to compare the samples obtained with different number of integration steps during sampling. 
Another indicator is the variance of $\nabla\Phi$ along the trajectories.

\paragraph{Hyperparameters}
As described in~\Cref{sec:implementation}, we select the width $w$ of the neural network, the number of time integration steps during training, the penalty parameters $\alpha_1, \alpha_2$, and the hyperparameters of the Adam optimizer (batch size and initial learning rate) from the values listed in~\Cref{tab:hyparam}. Following the original implementation of OT-Flow~\cite{onken2021ot}, we fix the network depth $K$ to 2. For more complex problems, we recommend exploring wider, deeper networks or more advanced architectures.

\section{Implementation and Experimental Setup}
\label{sec:implementation}
This section describes our implementations, experimental setups, and provides guidance for applying our techniques to new problems.

\paragraph{Implementation} The scripts for implementing our neural network approaches and running our numerical experiments are written in Python using PyTorch.
For datasets that are not publicly available, we provide the binary files we use in our experiments and the Python scripts for generating the data.
We have published the code and data along with detailed instructions on how to reproduce the results in our main repository, \texttt{\url{https://github.com/EmoryMLIP/PCP-Map.git}}.
Since COT-Flow is a generalization of a previous approach, we have created a fork for this paper at \texttt{\url{https://github.com/EmoryMLIP/COT-Flow.git}}. 

\paragraph{Hyperparameter selection}

Setting the values of key hyperparameters, describing both the network architecture and the optimization (training) algorithm, is crucial for achieving good performance from neural network approaches. The best values of such hyperparameters are typically both algorithm- and problem-specific. Ideally, one would like to have rigorous mathematical principles for choosing hyperparameters (e.g., akin to well-established methods for regularizing convex optimization problems in high-dimensional statistics~\cite{Negahban2012}), but this remains a largely open problem, outside the present scope. Usual practice is instead to tune hyperparameters via some black-box optimization method, such as Bayesian optimization~\cite{Snoek_2012}.

Rather than taking such an approach here, we use a simpler optimization approach that also better demonstrates the \emph{robustness} of our algorithms to sub-optimal hyperparameter values. We begin with the tensor product of candidate hyperparameter values given in \Cref{tab:hyparam} (with problem-specific modifications reported in \Cref{tab:tabhyparam} and \Cref{tab:swhyperparam}). From this candidate set, we \textit{randomly} select a smaller subset of {hyperparameter tuples} and perform pilot training runs for each tuple. (The size of this random subset is reported in each example below.) Pilot training consists of running the training procedures described in \Cref{sec:PCP-Map} or \Cref{sec:OTregularizedML} for a small number of epochs. The tuple of hyperparameter values achieving the best validation loss after pilot training is then selected for full training. Furthermore, we assess robustness by selecting the best $n$ (rather than single best) hyperparameter tuple after pilot training and performing full training with all $n$ of these setups; then we report the best, median, and worst values of the test loss (as in \Cref{tab:uci}).
We adopt this procedure as it represents a ``middle ground'' between very refined hyperparameter tuning and completely random selection.

\begin{table}[t]
    \caption{Hyperparameter selection, \Cref{sec:implementation}. Hyperparameter search spaces for PCP-Map and COT-Flow. Here $m$ denotes the size of the context feature or observation $\bfy$.} \label{tab:hyparam}
	\begin{center}
        \resizebox{.8\textwidth}{!}{
		\begin{tabular}{|l||c||c|}
			\hline
			Hyperparameters & PCP-Map & COT-Flow\\
			\hline\hline
			Batch size    & $\{2^5, 2^6, 2^7, 2^8\}$  & $\{2^5, 2^6, ..., 2^{10}\}$\\
			Learning rate   & $\{0.05,0.01 ,10^{-3}, 10^{-4}\}$ & $\{0.05, 0.01, 10^{-3}, 10^{-4}\}$ \\
			Feature Layer width, $w$ & $\{2^5, 2^6, ..., 2^9\}$ & $\{2^5, 2^6, ..., 2^{10}\}$\\
                Context Layer width, $u$ & $ \left\{\dfrac{w}{2^i}\bigg\vert \dfrac{w}{2^i} > m, i=0,1,...\right\}\cup \{m\}$ & \\
                Embedding feature width, $w_y$ & & $\{2^5, 2^6, 2^7\}$\\
                Embedding output width, $w_{yout}$ & & $\{2^5, 2^6, 2^7\}$\\
			Number of layers, $K$ & \{2, 3, 4, 5, 6\} & \{2\} \\
			Number of time steps, $n_t$ &  &   \{8, 16\}\\
			$[\log(\alpha_1), \log(\alpha_2)]$ & & [$\mathcal{U}$(-1, 3), $\mathcal{U}$(-1, 3)]\;\text{or}\\
            & & [$\mathcal{U}(2, 5)$, $\mathcal{U}(2, 5)$]\\
            [2pt]
			\hline
		\end{tabular}
        }
	\end{center}
\end{table}

\section{Numerical Experiments}
\label{sec:experiments}
We test the accuracy, robustness, efficiency, and scalability of our approaches from~\Cref{sec:PCP-Map,sec:OTregularizedML} using three problem settings that lead to different challenges and benchmark methods for comparison. All numerical experiments are conducted on one Quadro RTX 8000 GPU with 48 GB of RAM.
In~\Cref{sub:UCI}, we compare our proposed approaches to the Adaptive Transport Maps (ATM) approach developed in~\cite{baptista2022representation} %by repeating the experiments 
on estimating the joint and conditional distributions of six UCI tabular datasets.
In~\Cref{sec:stochLV}, we compare our approaches to a provably convergent approximate Bayesian computation (ABC) approach on accuracy and computational cost using the stochastic Lotka--Volterra model, which yields an intractable likelihood. Using this dataset, we also compare PCP-Map's and COT-Flow's sampling efficiency for different settings.
In~\Cref{sec:shallow}, we demonstrate the scalability of our approaches to higher-dimensional problems by comparing them to the flow-based neural posterior estimation (NPE) approach on an inference problem involving the 1D shallow water equations.
To demonstrate the improvements of PCP-Map over the amortized CP-Flow approach in the repository associated with ~\cite{huang2021convex}, we compare computational cost in~\Cref{sub:comp_cp_flow}.

\subsection{UCI Tabular Datasets}\label{sub:UCI}
We follow the experimental setup in \cite{baptista2022representation} by first removing the discrete-valued features and one variable of every pair with a Pearson correlation coefficient greater than 0.98.
We then partition the datasets into training, validation, and testing sets using an 8:1:1 split, followed by normalization.
For the joint and conditional tasks, we set $\bfx$ to be the second half of the features and the last feature, respectively. The conditioning variable $\bfy$ is set to be the remaining features for both tasks.

To perform joint density estimation and sampling, we use the inverse of the block-triangular generator in \eqref{eq: blocktriangular}
\begin{equation} \label{eq:map_inverse_Jacobian}
	T^{-1}(\bfx,\bfy) = 
	\begin{bmatrix*}[l]
            \mathsf{h}^{-1}(\bfy) \\   
            \gx^{-1}\left(\bfx; \bfy\right) 
	\end{bmatrix*}    
	\text{ and }\, 
	\nabla T^{-1}(\bfx,\bfy)
	= 
	\begin{bmatrix*}[l]
            \nabla_{\bfy} \mathsf{h}^{-1}(\bfy) & 0 \\
            \nabla_{\bfy} \gx^{-1}\left(\bfx; \bfy\right) & \nabla_{\bfx} \gx^{-1}\left(\bfx; \bfy\right)
	\end{bmatrix*}.
\end{equation}
We learn the map $T$ by minimizing the expected negative log-likelihood functional
\begin{equation}
	J_{{\rm NLL}}[T^{-1}] = \Ex_{\pi(\bfx,\bfy)} \left[ \hf \left\| T^{-1}(\bfx,\bfy) \right\|^2 - \log \det \nabla T^{-1}(\bfx,\bfy)  \right].
\end{equation}
Since the reference distribution, $\rho_\bfz = \mathcal{N}(0, I_{n+m})$, is of product type, we can decouple the objective functional into the following two terms
\begin{align*}
	J_{{\rm NLL}}[T^{-1}]  = J_{{\rm NLL},\bfx}[\gx^{-1}] + J_{{\rm NLL},\bfy}[\mathsf{h}^{-1}]
\end{align*}
with $J_{{\rm NLL},\bfx}$ defined in~\eqref{eq:JNNLx} and 
\begin{equation}
	J_{{\rm NLL},\bfy}[\mathsf{h}^{-1}] = 
	\mathbb{E}_{\bfy\sim \pi(\bfy)} \left[ \left\| \mathsf{h}^{-1}(\bfy) \right\|^2 - \log \det \nabla_\bfy \mathsf{h}^{-1}(\bfy)  \right].
\end{equation}

For PCP-Map, as proposed in \cite{pmlr-v70-amos17b}, we employ the gradient of a fully input convex neural network (FICNN) $\Finv_{\theta_F}:\mathbb{R}^n\to \mathbb{R}$, parameterized by weights $\theta_F$, to represent the inverse generator $\mathsf{h}^{-1}$. A general $K$-layer FICNN can be expressed as the following sequence starting with $\bfs_0 = \bfy$:
\begin{equation}\label{eq:FICNN}
    \bfs_{k+1} = \sigma^{(s)}\left( \bfL^{(w)}_k \bfs_k + \bfL^{(y)}_k\bfy + \bfb_k \right)
\end{equation}
for $k = 0, \ldots, K-1$. Here $\sigma^{(s)}$ is the softplus activation function. Since the FICNN map is not part of our contribution, we use the input-augmented ICNN implementation used in \cite{huang2021convex}.
For the second map component $\gx^{-1}$, we use the gradient of a PICNN as described in \Cref{sec:PCP-Map}.
For COT-Flow, we construct two distinct neural network parameterized potentials $\Phi_x$ and $\Phi_y$. Here, $\Phi_y$ only takes the conditioning variable $\bfy$ as inputs and can be constructed exactly like in \cite{onken2021ot}. $\Phi_x$ acts on all features and is the same potential described in \Cref{sec:OTregularizedML}. 
We learn the inverse conditional generator $g^{-1}_\theta$ to perform conditional density estimation. This is equivalent to learning only the weights $\theta$ of the PICNN $\Ginv_\theta$ for PCP-Map. For COT-Flow, we only construct and learn the potential $\Phi_x$. 

\begin{table}[t]
    \caption{UCI tabular datasets experiment, \Cref{sub:UCI}. Hyperparameter search space.} \label{tab:tabhyparam}
	\begin{center}
        \resizebox{.8\textwidth}{!}{
		\begin{tabular}{|l||c||c|}
			\hline
			Hyperparameters & PCP-Map & COT-Flow\\
			\hline\hline
			Batch size    & \{32, 64\}  & \{32, 64\}\\
			Learning rate   & \{0.01, 0.005, 0.001\} & \{0.01, 0.005, 0.001\}  \\
			Feature Layer width, $w$ & \{32, 64, 128, 256, 512\} & \{32, 64, 128, 256, 512\}\\
                Context Layer width, $u$ & \{$\frac{w}{2^i}| \frac{w}{2^i} > m, i=0,1,...$\}$\cup$\{$m$\} & \\
			Number of layers, $K$ & \{2, 3, 4, 5, 6\} & \{2\} \\
			Number of time steps, $n_t$ &  &   \{8, 16\}\\
			$[\log(\alpha_1), \log(\alpha_2)]$ & & [$\mathcal{U}$(-1, 3), $\mathcal{U}$(-1, 3)]\\[2pt]
			\hline
		\end{tabular}
        }
	\end{center}
\end{table}

The hyperparameter search space we use for this experiment is presented in \Cref{tab:tabhyparam}. We select smaller batch sizes and large learning rates as this leads to fast convergence on these relatively simple problems. For each dataset, we select the ten best hyperparameter combinations based on a pilot run for full training and report the best, median, and worst results. For PCP-Map's pilot run, we performed 15 epochs for all three conditional datasets, three epochs for the Parkinson’s and the White Wine datasets, and four epochs for Red Wine using 100 randomly sampled combinations. 
For COT-Flow, we limit the pilot runs to only 50 sampled combinations due to a narrower sample space on model architecture. 
To assess the robustness of our approaches, we performed five full training runs with random initializations of network weights for each of the ten hyperparameter combinations for each dataset.

In \Cref{tab:UCI-results}, we present the best, median, and worst test-time mean NLL as well as the maximum mean discrepancy (MMD) between generated and true samples. For two sample sets  $P=\{\bfp_i\}^N_{i=1}$ and $Q=\{\bfq_i\}^M_{i=1}$, the MMD is computed as:
\begin{eqnarray}\label{eq: emp_mmd}
\text{MMD}(P,Q) = & \frac{1}{N^2}\sum_{i=1}^{N}\sum_{j=1}^{N} k(\bfp_i,\bfp_j)
+ \frac{1}{M^2}\sum_{i=1}^{M}\sum_{j=1}^{M} k(\bfq_i,\bfq_j) \\ & - \frac{2}{NM}\sum_{i=1}^{N}\sum_{j=1}^{M} k(\bfp_i,\bfq_j), \nonumber
\end{eqnarray}
with $k$ chosen to be a squared exponential kernel with bandwidth one, following~\cite{Gretton_2012,Peyre_2019}. 
Results are reported with empirical standard deviations obtained by repeating training with five random neural network weight initializations (for each fixed set of hyperparameter values). We run our two proposed approaches on six UCI datasets, and compare to the best reported NLL for ATM.
The table demonstrates that the best models outperform ATM for all datasets and that the median performance is typically superior. 
Overall, COT-Flow slightly outperforms PCP-Map in terms of NLL, while the opposite is true in terms of MMD for the best models.
The improvements are more pronounced for the conditional density estimation tasks, where even the worst hyperparameters from PCP-Map and COT-Flow improve over ATM by a substantial margin. 

\begin{table}[t]\label{tab:uci}
	\centering
    \footnotesize
    \caption{UCI tabular datasets experiment, \Cref{sub:UCI}. Mean NLL and MMD comparisons on test data, for ATM, PCP-Map, and COT-Flow. For each approach, we report the best, median, and worst results across five training runs and various hyperparameter combinations, with standard deviations shown ($\pm$). Lower values are better, and the best results are highlighted in bold.}    \label{tab:UCI-results}
    \resizebox{0.96\columnwidth}{!}{%
	\begin{tabular}{@{}|@{\,}l@{\,}||c|c|c|c|c|c@{\,}|@{}}
		\hline
		& \multicolumn{3}{|c|}{joint} & \multicolumn{3}{|c|}{conditional}\\ \hline
		dataset name               & Parkinson's & white wine & red wine & concrete & energy & yacht \\
		dimensionality             & $d=15$      & $d=11$     & $d=11$   & $d=9$    & $d=10$ &   $d=7$ \\ 
		\# samples                & $N=5875$    & $N=4898$   & $N=1599$ & $N=1030$ & $N=768$&   $N=308$ \\
        \hline\hline
        \multicolumn{7}{|c|}{Negative log-likelihood (NLL)} \\[1pt] \hline
		ATM~\cite{baptista2022representation} & $2.8\pm0.4$ & $11.0\pm0.2$ & $9.8\pm0.4$ & $3.1\pm0.1$ & $1.5\pm0.1$ & $0.5\pm0.2 $    
		\\ 
		PCP-Map (best)                        & 1.59$\pm$0.08 & 10.81$\pm$0.15 & 8.80$\pm$0.11 & 0.19$\pm$0.14 & -1.15$\pm$0.08 & -2.76$\pm$0.18    \\ 
		PCP-Map (median)                      & 1.96$\pm$0.08 & 10.99$\pm$0.24 & 9.90$\pm$0.53 & 0.28$\pm$0.07 & -1.02$\pm$0.16 & -2.42$\pm$0.25    \\ 
		PCP-Map (worst)                       & 2.34$\pm$0.09 & 12.53$\pm$2.68 & 11.08$\pm$0.72 & 1.18$\pm$0.54 & 0.30$\pm$0.63 & -0.27$\pm$1.28     \\ 
		COT-Flow (best)                        & \bf{1.58$\pm$0.09} & \bf{10.45$\pm$0.08} & \bf{8.54$\pm$0.13} & \bf{0.15$\pm$0.05} & \bf{-1.19$\pm$0.09} & \bf{-3.14$\pm$0.14}          \\ 
		COT-Flow (median)                      & 2.72$\pm$0.34 & 10.73$\pm$0.05 & 8.71$\pm$0.12 & 0.21$\pm$0.04 & -0.83$\pm$0.05 & -2.77$\pm$0.12          \\ 
		COT-Flow (worst)                       & 3.27$\pm$0.15 & 11.04$\pm$0.28 & 9.00$\pm$0.05 & 0.35$\pm$0.04 & -0.56$\pm$0.04 & -2.38$\pm$0.11          \\[3pt] \hline\hline
        \multicolumn{7}{|c|}{Maximum mean discrepancy (MMD) $\times 10^{-2}$} \\[1pt] \hline
        PCP-Map (best)                        & \bf{3.3$\pm$0.3} & \bf{2.8$\pm$0.1} & 4.3$\pm$0.3 & \bf{2.7$\pm$1.1} & \bf{1.4$\pm$0.4} & \bf{0.7$\pm$0.1}    \\ 
		PCP-Map (median)                      & 3.5$\pm$0.4 & 3.1$\pm$0.3 & 4.7$\pm$0.3 & 4.0$\pm$1.1 & 1.9$\pm$0.5 & 1.2$\pm$0.5    \\ 
		PCP-Map (worst)                       & 4.0$\pm$0.6 & 3.3$\pm$0.2 & 5.5$\pm$0.3 & 4.8$\pm$1.0 & 3.8$\pm$1.7 & 1.7$\pm$0.6     \\ 
		COT-Flow (best)                        & 3.4$\pm$0.3 & 3.0$\pm$0.1 & \bf{4.2$\pm$0.1} & 2.9$\pm$1.4 & 1.5$\pm$0.3 & 1.0$\pm$0.3          \\ 
		COT-Flow (median)                      & 4.1$\pm$0.4 & 3.2$\pm$0.1 & 4.6$\pm$0.3 & 4.3$\pm$0.9 & 3.1$\pm$1.4 & 1.6$\pm$0.7          \\ 
		COT-Flow (worst)                       & 4.8$\pm$0.2 & 4.4$\pm$0.1 & 6.7$\pm$0.3 & 7.1$\pm$3.0 & 7.4$\pm$1.9 & 2.7$\pm$0.3          \\ \hline
	\end{tabular}%
	}
\end{table}

\subsection{Stochastic Lotka--Volterra} \label{sec:stochLV}
We compare our approaches to an ABC approach based on sequential Monte Carlo (SMC) for likelihood-free Bayesian inference using the stochastic Lotka--Volterra (LV) model~\cite{wilkinson2018stochastic}. The LV model is a stochastic process whose dynamics describe the evolution of the populations $S(t) = (S_1(t),S_2(t))$ of two interacting species, e.g., predators and prey. These populations start from a fixed initial condition $S(0) = (50,100)$. The parameter $\bfx \in \R^4$ determines the rate of change of the populations over time, and the observation $\bfy \in \R^9$ contains summary statistics of the time series generated by the model. This results in an observation vector with nine entries: the mean, the log-variance, the auto-correlation with lags one and two, and the cross-correlation coefficient. The procedure for sampling a trajectory of the species populations is known as Gillespie’s algorithm.

Given a prior distribution for the parameter $\bfx$, we aim to sample from the posterior distribution corresponding to an observation $\bfy^*$. As in~\cite{NIPS2016_6aca9700}, we consider a log-uniform prior distribution for the parameters whose density (of each component) is given by 
$\pi(\log x_i) = \mathcal{U}(-5, 2)$.
As a result of the stochasticity that enters non-linearly in the dynamics, the likelihood function is not available in closed form. Hence, this model is a popular benchmark for likelihood-free inference algorithms as they avoid evaluating $\pi(\bfy|\bfx)$~\cite{NIPS2016_6aca9700}. 

% Describe TC and OT set up
We generate two training sets consisting of 50k and 500k samples from the joint distribution $\pi(\bfx, \bfy)$ obtained using Gillespie’s algorithm. 
To account for the strict positivity of the parameter, which follows a log-uniform prior distribution, we perform a log transformation of the $\bfx$ samples. This ensures that the conditional distribution of interest has full support, which is needed to find a diffeomorphic map to a Gaussian.
We split the log-transformed data into ten folds and use nine folds of the samples as training data and one fold as validation data. 
We normalize the training and validation sets using the training set's empirical mean and standard deviation. 

\begin{figure}[t]
\centering
\includegraphics[width=\linewidth]{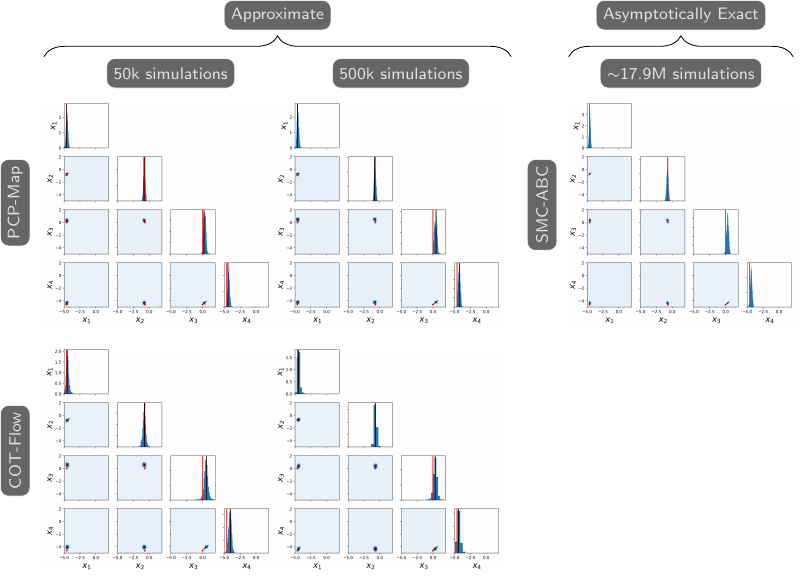}
\caption{Stochastic Lotka--Volterra experiment, \Cref{sec:stochLV}. Comparison of estimated posterior distributions and MAP points between the proposed approaches and SMC-ABC ($\bfx^* = (0.01, 0.5, 1, 0.01)^\top$). Red dots/bars represent $\bfx^\ast$; black crosses/bars represent MAP points. \textbf{Column 1}: Proposed approaches (50k training samples). \textbf{Column 2}: Proposed approaches (500k training samples). \textbf{Column 3}: SMC-ABC (around 17.9 million simulations).}
\label{fig:NNvsABC}
\end{figure}

For the pilot run, we use the same sample space in \Cref{tab:tabhyparam} except expanding the batch size space to \{64, 128, 256\} for PCP-Map and \{32, 64, 128, 256\} for COT-Flow to account for the increase in sample size. We also fixed, for PCP-Map, $w = u$.
During PCP-Map's pilot run, we perform two training epochs with 100 hyperparameter combination samples using the 50k dataset.
For COT-Flow, we only perform one epoch of pilot training as it is empirically observed to be sufficient. 
We then use the best hyperparameter combinations to train our models on the 50k and 500k datasets to learn the posterior for the normalized parameter in the log-domain. 
After learning the maps, we used their inverses, the training data mean and standard deviation, and the log transformations to yield parameter samples in the original domain.

The SMC-ABC algorithm finds parameters that match the observations with respect to a selected distance function by gradually reducing a tolerance $\epsilon > 0$, which leads to samples from the true posterior exactly as $\epsilon \rightarrow 0$. For our experiment, we consider inference problems arising from observations associated with two true parameters. We allow $\epsilon$ to converge to $0.1$ for the first true parameter and $0.15$ for the second. 

We evaluate our approaches for maximum-a-posteriori (MAP) estimation and posterior sampling. We consider a true parameter 
$\bfx^* = (0.01, 0.5, 1, 0.01)^\top$, which was carefully chosen to give rise to oscillatory behavior in the population time series. Given one observation $\bfy^* \sim \pi(\bfy | \bfx^*)$, we first identify the MAP point by maximizing the estimated log-likelihoods provided by our approaches. Then, we generate 2000 samples from the approximate posterior $\pi(\bfx | \bfy^*)$ using our approaches. \Cref{fig:NNvsABC} presents one and two-dimensional marginal histograms and scatter plots of the MAP point and samples, compared against 2000 samples generated by the SMC-ABC algorithm from~\cite{bonassi2015sequential}. %\rbtodo{It would be great to list the number of samples used in ABC and sampled from the generator in the plots below. \textcolor{red}{Updated!}}. 
Our approaches yield samples tightly concentrated around the MAP points that are close to the true parameter $\bfx^*$.

\begin{figure}[t]
\centering
\includegraphics[width=\linewidth]{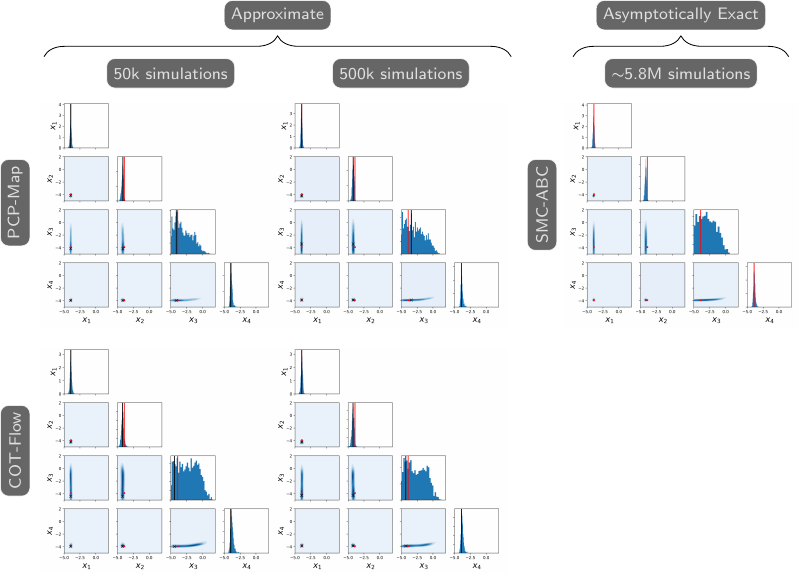}
\caption{Stochastic Lotka--Volterra experiment, \Cref{sec:stochLV}. Comparison of estimated posterior distributions and MAP points between the proposed approaches and SMC-ABC ($\bfx^* = (0.02, 0.02, 0.02, 0.02)^\top$). Red dots/bars represent $\bfx^\ast$; black crosses/bars represent MAP points. \textbf{Column 1}: Proposed approaches (50k training samples). \textbf{Column 2}: Proposed approaches (500k training samples). \textbf{Column 3}: SMC-ABC (around 5.8 million simulations).}
\label{fig:NNvsABCNew}
\end{figure}

To provide more evidence that our learning approaches indeed solve the amortized problem, \Cref{fig:NNvsABCNew} shows the MAP points and approximate posterior samples generated from a new random observation $\bfy^*$ corresponding to the true parameter $\bfx^* = (0.02, 0.02, 0.02, 0.02)^\top$. We observe similar concentrations of the MAP point and posterior samples around the true parameter and similar correlations learned by the generative model and ABC, for example, between the third and other parameters. 

Efficiency-wise, PCP-Map and COT-Flow yield similar approximations to ABC at a fraction of the computational cost of ABC. The latter requires approximately 5 or 18 million model simulations for each conditioning observation, while the learned approaches use the same $50$ thousand simulations to amortize over the observation. These savings generally offset the hyperparameter search and training time for the proposed approaches, which is typically less than half an hour per full training run on a GPU. For some comparison, SMC-ABC took 15 days to reach $\epsilon=0.1$ for $\bfx^*=(0.01, 0.5, 1, 0.01)^\top$.

\begin{figure}[t]
\centering
\includegraphics[width=0.7\linewidth]{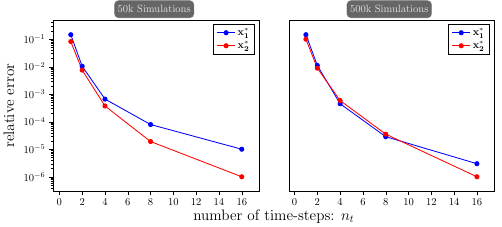}
\caption{Stochastic Lotka--Volterra experiment, \Cref{sec:stochLV}. Relative normed errors between COT-Flow posterior samples generated with smaller number of time-steps $n_t$ and samples generated with $n_t=32$. Here $\bfx^*_1 = (0.01, 0.5, 1, 0.01)^\top$ and $\bfx^*_2 = (0.02, 0.02, 0.02, 0.02)^\top$.}
\label{fig:err-diff-nt}
\end{figure}

In the above experiments, we used $n_t=32$ to generate posterior samples during testing for the COT-Flow. 
One can also decrease $n_t$ to generate samples faster without sacrificing much accuracy, as shown in \Cref{fig:err-diff-nt} and \Cref{tab:cotvspcp}. To establish such a comparison, we increase the $l$-BFGS tolerance when sampling using PCP-Map, which produces a similar effect on sampling accuracy than decreasing the number of time steps for COT-Flow. 

\begin{table}[t] \label{tab:cotvspcp}
    \caption{Stochastic Lotka--Volterra experiment, \Cref{sec:stochLV}. Sampling efficiency comparisons between PCP-Map and COT-Flow in terms of GPU time in seconds (s). We report the mean and standard deviation over five runs, respectively.} \label{tab:desamplecomp}
	\begin{center}
		\begin{tabular}{||c||c||}
			\hline
			COT-Flow Sampling (s) & PCP-Map Sampling (s)\\
			\hline\hline
			$n_t$=32\quad\quad 0.538 $\pm$ 0.003 & tol=1e-6 \quad\quad0.069 $\pm$ 0.004\\
                $n_t$=16\quad\quad 0.187 $\pm$ 0.004 & tol=1e-5 \quad\quad0.066 $\pm$ 0.004\\
                $n_t$=8\quad\quad 0.044 $\pm$ 0.000 & tol=1e-4 \quad\quad0.066 $\pm$ 0.004\\
                $n_t$=4\quad\quad 0.025 $\pm$ 0.001 & tol=1e-3 \quad\quad0.066 $\pm$ 0.004\\
                $n_t$=2\quad\quad 0.012 $\pm$ 0.000 & tol=1e-2 \quad\quad0.059 $\pm$ 0.001\\
                $n_t$=1\quad\quad 0.006 $\pm$ 0.000 & tol=1e-1 \quad\quad0.049 $\pm$ 0.004\\
                \hline
		\end{tabular}
	\end{center}
\end{table}

\subsection{1D Shallow Water Equations} \label{sec:shallow}
The shallow water equations model wave propagation through shallow basins described by the depth profiles parameter, $\bfx\in\R^{100}$, which is discretized at 100 equidistant points in space. 
After solving the equations over a time grid with 100 cells, the resulting wave amplitudes form the 10k-dimensional raw observations. 
As in \cite{GATSBI}, we perform a 2D Fourier transform on the raw observations and concatenate the real and imaginary parts since the waves are periodic. 
We define this simulation-then-Fourier-transform process as our forward model and denote it as $\Psi(\bfx)$. 
Additive Gaussian noise is introduced into the output of $\Psi$, giving us the observations $\bfy = \Psi(\bfx) + 0.25\bfepsilon$, where  $\bfy,\bfepsilon \in\R^{200\times 100}$ and $\bfepsilon_{i, j}\sim \mathcal{N}(0, 1)$.
We aim to use the proposed approaches to learn the posterior $\pi(\bfx|\bfy)$. 

We follow instructions from \cite{GATSBI} to set up the experiment and obtain 100k samples from the joint distribution $\pi(\bfx, \bfy)$ as the training dataset using the provided scripts. We use the prior distribution defined as
\begin{equation*}
     \pi(\bfx) = \mathcal{N}(10\cdot\mu\mathbf{1}_{100}, \bfSigma) \; \rm \; with \;\; \bfSigma_{ij}=\sigma\exp\left( \frac{-(i-j)^2}{2\tau} \right), \;\sigma=15, \;\tau=100.
\end{equation*}
Using a principal component analysis (PCA), we analyze the intrinsic dimensions of $\bfx$ and $\bfy$.
To ensure that the large additive noise $0.25\bfepsilon$ does not affect our analysis, we first study the noise-free prior predictive by collecting a set of 100k forward model evaluations and forming the estimated covariance $\Cov(\Psi(\bfX)) \approx \frac{1}{N-1}\Psi(\bfX)^\top \Psi(\bfX)$. Here, $\Psi(\bfX) \in \R^{100000 \times 20000}$ stores the forward model outputs row-wise in a matrix. 
The top 3500 modes explain around $96.5\%$ of the variance. 
A similar analysis on the noise-present training dataset shows that the top 3500 modes, in this case, only explain around $75.6\%$ of the variance due to the added noise. 
To address the rank deficiency, we construct a projection matrix $\bfV_{\rm proj}$ using the top 3500  eigenvectors of $\Cov(\Psi(\bfX))$ and obtain the projected observations $\bfy_{\rm proj} = \bfV_{\rm proj}^\top\bfy$ from the training datasets.
We then perform a similar analysis for $\bfx$ and discovered that the top 14 modes explained around $99.9\%$ of the variance. 
Hence, we obtain $\bfx_{\rm proj} \in \R^{14}$ as the projected parameters.

We then trained our approaches to learn the reduced posterior, $\pi_{\rm proj}(\bfx_{\rm proj}|\bfy_{\rm proj})$. For comparisons, we trained the flow-based NPE approach to learn the same posterior by following~\cite{GATSBI}.
For COT-Flow, we add a 3-layer fully connected neural network with $\tanh$ activation to embed $\bfy_{\rm proj}$.
To pre-process the training dataset, we randomly select 5\% of the dataset as the validation set and use the rest as the training set. 
We project both $\bfx$ and $\bfy$ and then normalize them by subtracting the empirical mean and dividing by the empirical standard deviations of the training data.

We employ the sample space presented in \Cref{tab:swhyperparam} for the pilot runs.
\begin{table}[t]
    \caption{Shallow water experiment, \Cref{sec:shallow}. Hyperparameter search space for the 1D shallow water experiment.} \label{tab:swhyperparam}
	\begin{center}
            \resizebox{.8\textwidth}{!}{
		\begin{tabular}{|l||c||c|}
			\hline
			Hyperparameters & PCP-Map & COT-Flow\\
			\hline\hline
			Batch size    & \{64, 128, 256\}  & $\{2^7, 2^8, 2^9, 2^{10}\}$\\
			Learning rate   & \{$10^{-2}, 10^{-3}, 10^{-4}$\} & $\{10^{-2}, 10^{-3}, 10^{-4}\}$ \\
			Feature layer width, $w$ & \{32, 64, 128, 256, 512\} & $\{512, 1024\}$ \\
                Context layer width, $u$ & \{$u = w$\} & \\
                Embedding feature width, $w_y$ & & $\{2^5, 2^6, 2^7\}$\\
                Embedding output width, $w_{yout}$ & & $\{2^5, 2^6, 2^7\}$\\
			Number of layers, $K$ & \{2, 3, 4, 5, 6\} & \{2\} \\
			Number of time steps, $n_t$ &  &   \{8, 16\}\\
			$[\log(\alpha_1), \log(\alpha_2)]$ & & [$\mathcal{U}(2, 5)$, $\mathcal{U}(2, 5)$]\\[2pt]
			\hline
		\end{tabular}
        }
	\end{center}
\end{table}
For COT-Flow, we select a $w$ sample space with larger values for maximum expressiveness and allow multiple optimization steps over one batch, randomly selected from $\{8, 16\}$.
We then use the best hyperparameter combination based on the validation loss for the full training. For NPE, we used the \texttt{sbi} package~\cite{tejero-cantero2020sbi} and the scripts provided by \cite{GATSBI}.

\begin{figure}[t]
\centering
\includegraphics[width=0.95\linewidth]{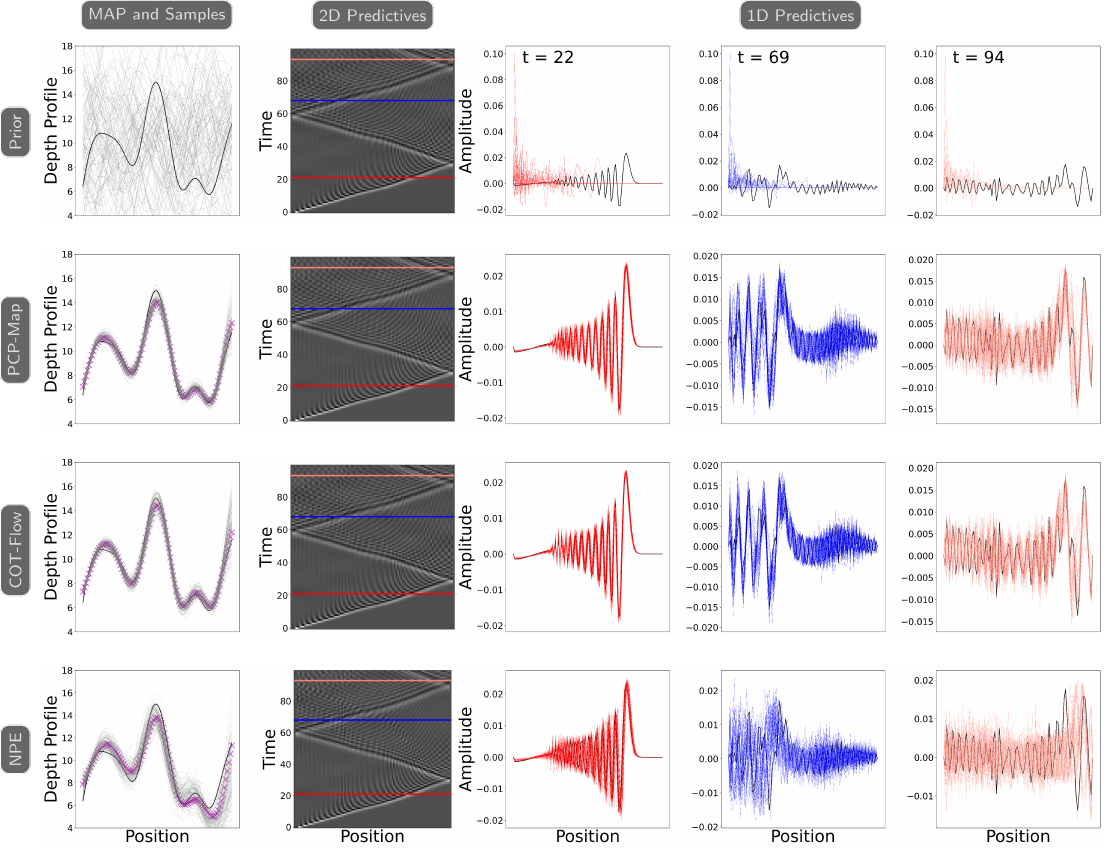}
\caption{Shallow water experiment, \Cref{sec:shallow}. Comparison of MAP estimation accuracy, posterior sampling quality, and posterior predictive accuracy among PCP-Map, COT-Flow, and NPE.  
\textbf{Column 1}: Ground truth parameter $\bfx^\ast$ (black), prior parameter samples (gray, row 1), generated posterior samples (gray, row 2-4), and MAP points (purple cross, row 2-4). 
\textbf{Column 2}: 2D wave simulation images over 100 time and space grids. Horizontal color lines mark 3 time slices shown in Columns 3–5. Ground truth simulation (row 1) and simulation from generated posterior samples (row 2-4).
\textbf{Columns 3--5}: Wave amplitudes across the 3 time slices. Ground truth simulations (black), simulations from prior samples (row 1) and generated posterior samples (rows 2–4).}
\label{fig:cot_pcp_sw}
\end{figure}

We first compare the accuracy of the MAP points, posterior samples, and posterior predictives across the three approaches. The MAP points are obtained using the same method as in \Cref{sec:stochLV}.
For posterior sampling, we first sample a ``ground truth" $\bfx^* \sim\pi(\bfx)$ and obtain the associated ground truth reduced observation $\bfy^*_{\rm proj} = \bfV_{\rm proj}^\top(\Psi(\bfx^*) + 0.25\bfepsilon)$. 
Then, we use the three approaches to sample from the posterior $\pi_{\rm proj}(\bfx_{\rm proj} | \bfy^*_{\rm proj})$.
This allows us to obtain approximate samples $\bfx\sim \pi(\bfx | \bfy^*)$. The posterior predictives are obtained by solving the forward model for the generated parameters. Through \Cref{fig:cot_pcp_sw}, we observe that the MAP points, posterior samples, and predictives produced by PCP-Map and COT-Flow are more concentrated around the ground truth than those produced by NPE. 

\begin{figure}[t]
\centering
\includegraphics[width=0.32\linewidth]{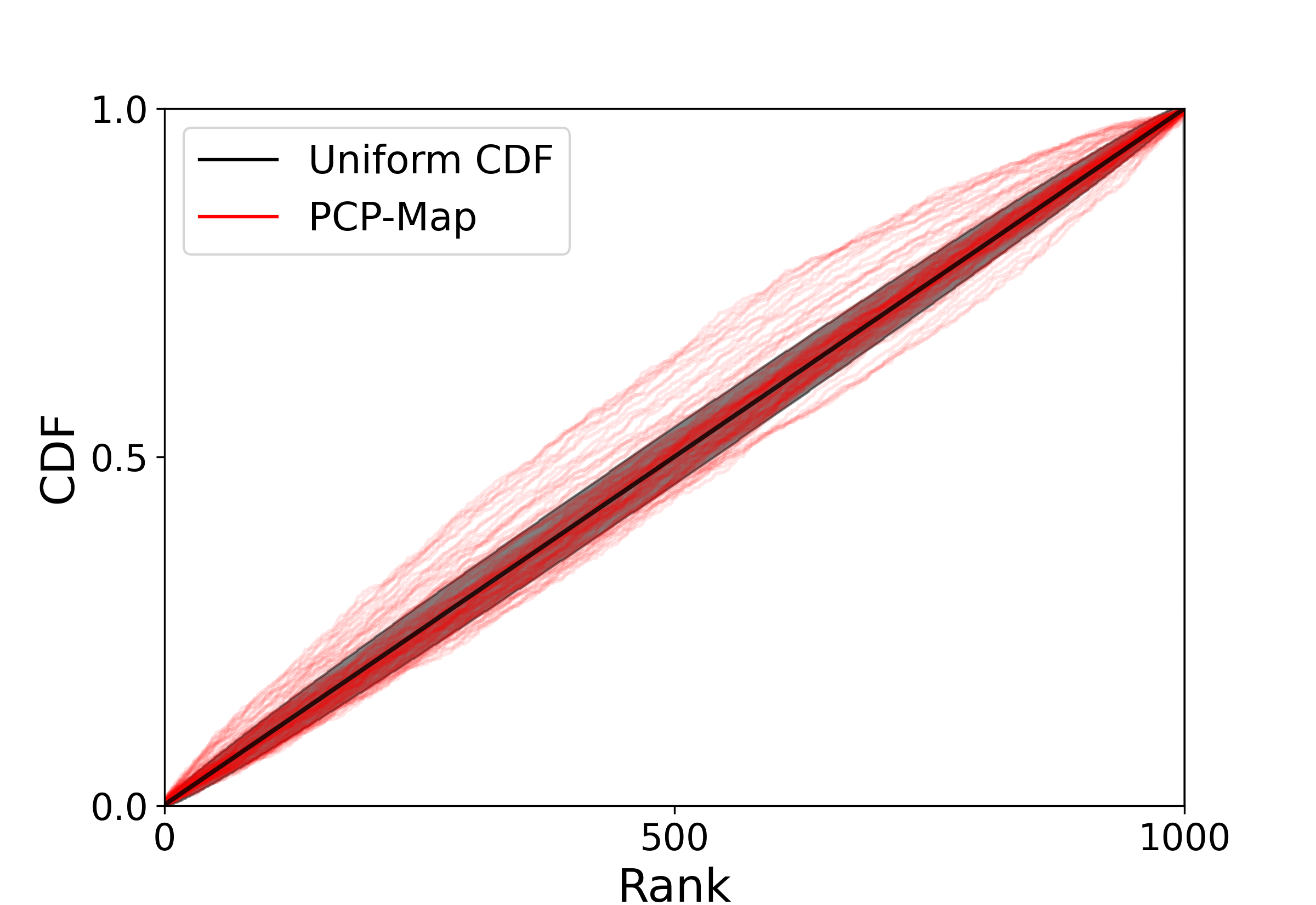}
\includegraphics[width=0.32\linewidth]{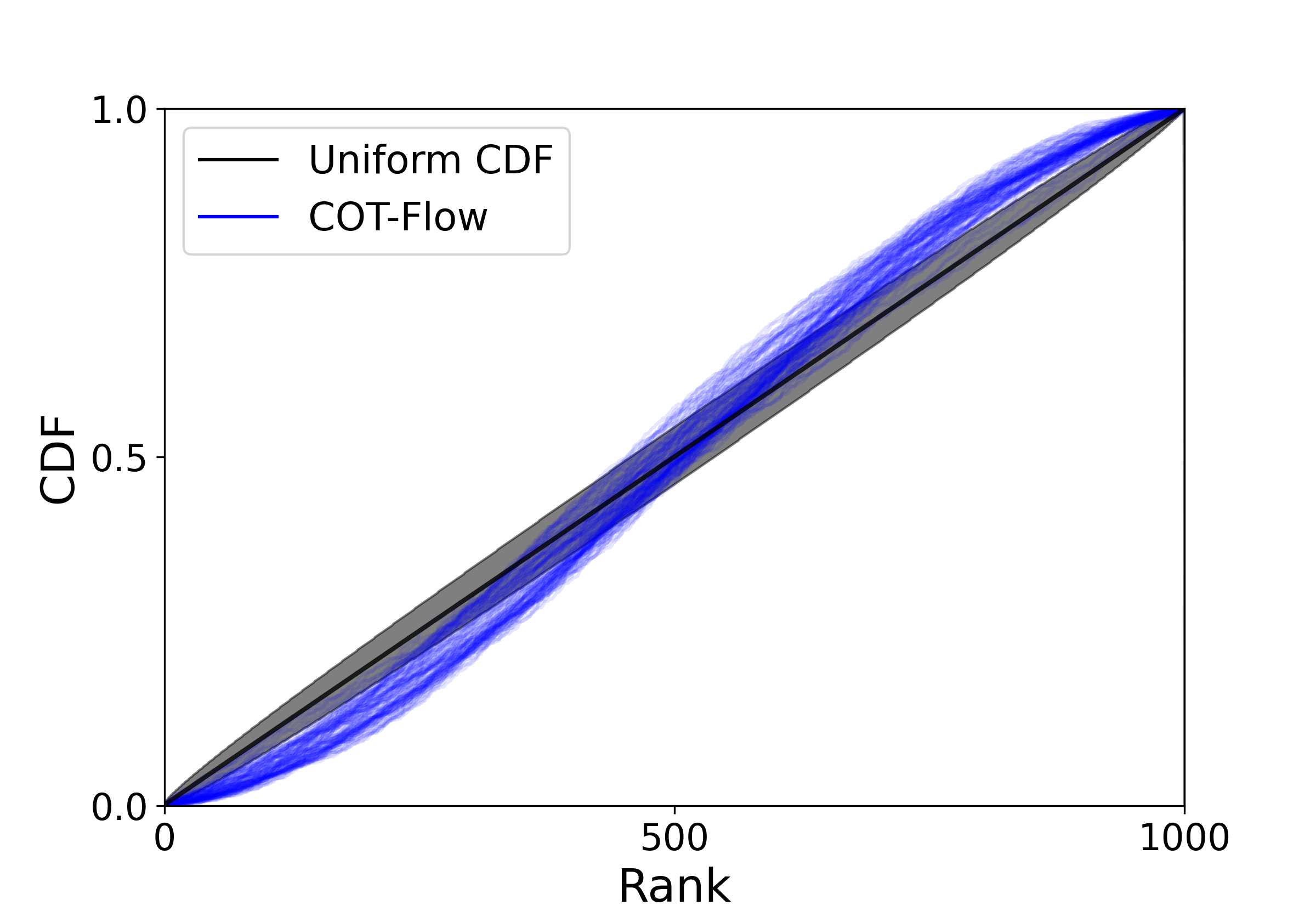}
\includegraphics[width=0.32\linewidth]{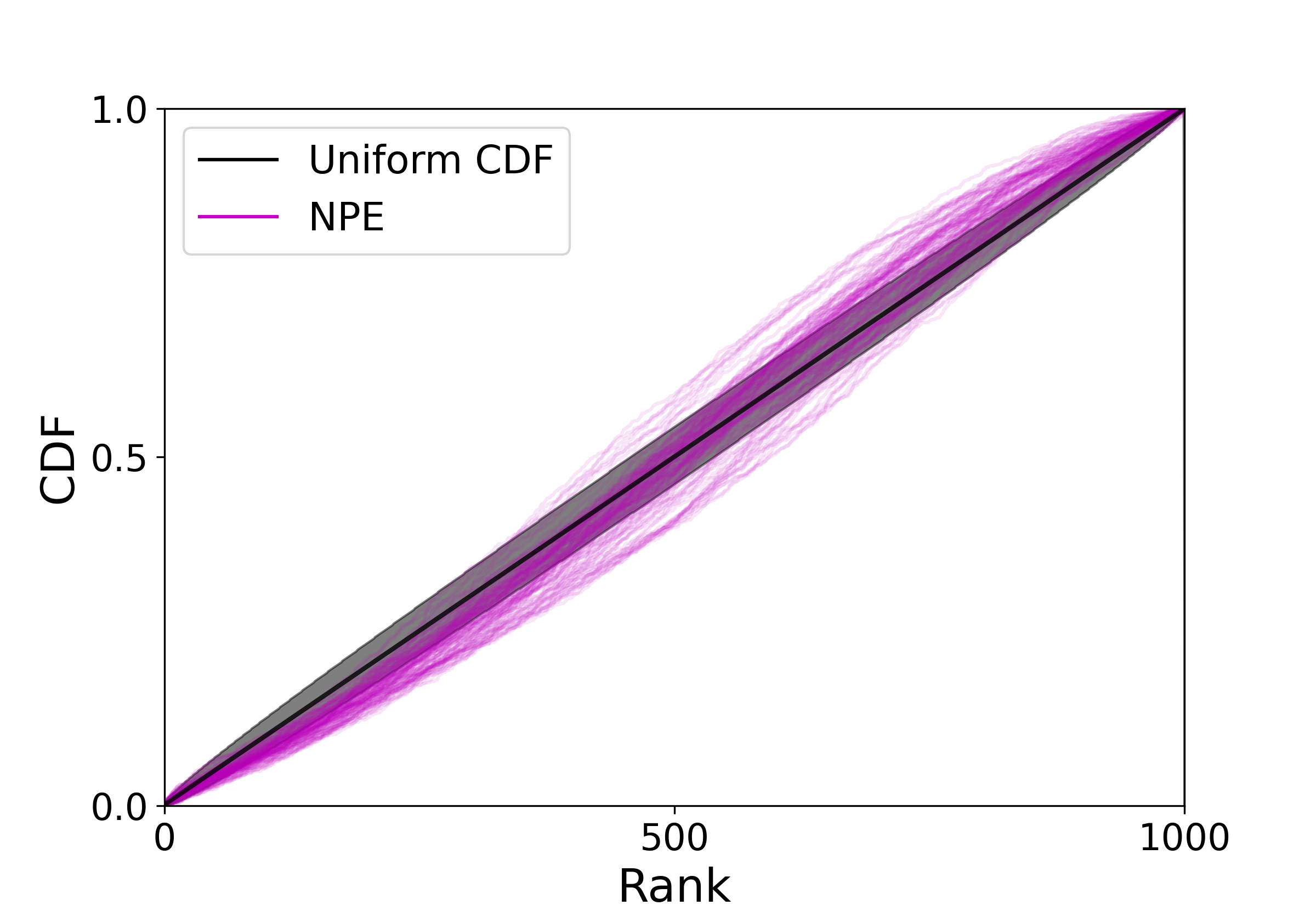}
\caption{ Shallow water experiment, \Cref{sec:shallow}. SBC analysis for PCP-Map, COT-Flow, and NPE. Each colored line represents the empirical cumulative distribution function (CDF) of the SBC rank associated with one posterior sample dimension. The more the empirical CDF resembles the CDF of the uniform distribution (i.e., the identity function), the better.}
\label{fig:sw_sbc}
\end{figure}

We perform the simulation-based calibration (SBC) analysis described in ~\cite[App. D.2]{GATSBI} to further assess the three approaches' accuracy; see \Cref{fig:sw_sbc}. We can see that, while they are all well calibrated, the cumulative distribution functions of the rank statistics produced by PCP-Map align almost perfectly with the CDF of a uniform distribution besides a few outliers. 

\begin{figure}[t]
\centering
\includegraphics[width=0.8\linewidth]{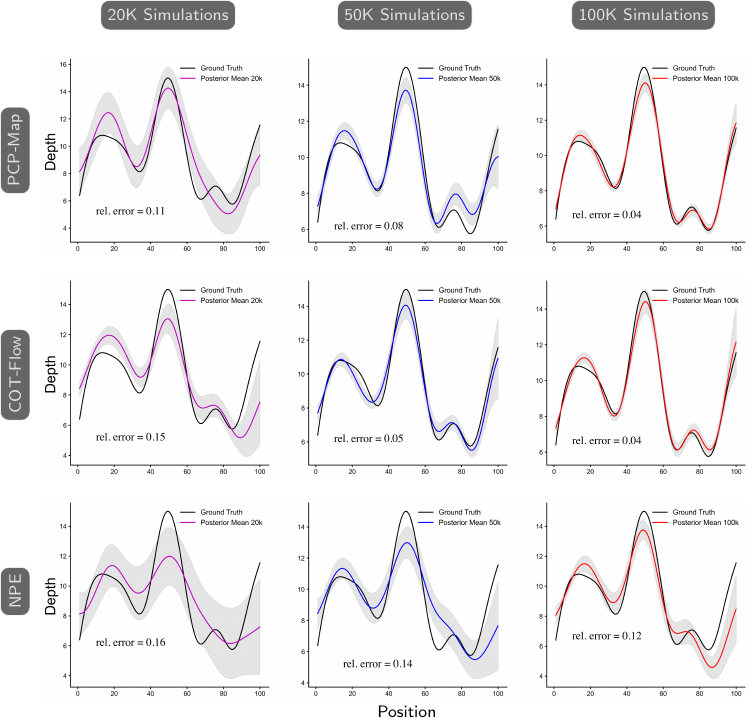}
\caption{Shallow water experiment, \Cref{sec:shallow}. Posterior sample quality comparison between PCP-Map, COT-Flow, and NPE trained on 20k, 50k, and 100k samples. Each plot shows the ground truth $\bfx^\ast$ (black), posterior mean (color), and standard deviation (gray band) over 100 samples.}
\label{fig:num_sims}
\end{figure}

Finally, we analyze the three approaches' efficiency in terms of the number of forward model evaluations. We train the models using the best hyperparameter combinations from the pilot run on two extra datasets with 50k and 20k samples. 
We compare the posterior samples' mean and standard deviation against $\bfx^*$ across three approaches trained on the 100k, 50k, and 20k sized datasets as presented in \Cref{fig:num_sims}. 
We see that PCP-Map and COT-Flow can generate posterior samples centered more closely around the ground truth than NPE using only 50k training samples, which translates to higher computational efficiency since fewer forward model evaluations are required.

\subsection{Comparing PCP-Map to Amortized CP Flow}\label{sub:comp_cp_flow}
We conduct this comparative experiment using the shallow water equations problem for its high dimensionality.
We include the more challenging task of learning $\pi(\bfx|\bfy_{\rm proj})$, obtained without projecting the parameter, to test the two approaches most effectively.
We followed \cite{huang2021convex} and its associated GitHub repository as closely as possible to implement the amortized CP-Flow.
To ensure as fair of a comparison as possible, we used the hyperparameter combination from the amortized CP-Flow pilot run for learning $\pi(\bfx|\bfy_{\rm proj})$ and the combination from the PCP-Map pilot run for learning $\pi_{\rm proj}(\bfx_{\rm proj}|\bfy_{\rm proj})$. 
Note that for learning $\pi(\bfx|\bfy_{\rm proj})$, we limited the learning rate to 0.0001 for which we observed reasonable convergence. 

In the experiment, we observed that amortized CP-Flow's Hessian vector product function gave NaNs consistently when computing the stochastic log-determinant estimation. 
Thus, we resorted to exact computation for the pilot and training runs. PCP-Map circumvents this as it uses exact log-determinant computations.
Each model was trained five times to capture possible variance. 

For comparison, we use the exact mean negative log-likelihood as the metric for accuracy and GPU time as the metric for computational cost. 
We only record the time for loss function evaluations and optimizer steps for both approaches. The comparative results are presented in \Cref{tab:cpvspcp}. 
We see that PCP-Map and amortized CP-Flow reach similar training and validation accuracy measured by NLL. 
However, PCP-Map on average takes roughly 7 and 3 times less GPU time, respectively, to achieve that accuracy than amortized CP-Flow. 
Possible reasons for the increased efficiency of PCP-Map's are its use of ReLU non-negative projection, exact and vectorized Hessian computation, removal of activation normalization, and gradient clipping.

\begin{table}[t] \label{tab:cpvspcp}
    \caption{Computational cost comparison (\Cref{sub:comp_cp_flow}) between amortized CP-Flow and PCP-Map. We report the mean and standard deviation of GPU time (s), training/validation NLL, and number of model parameters $M$ over five runs. The better result is highlighted in bold.}\label{tab:timecomp}
	\begin{center}
            \resizebox{.93\textwidth}{!}{
		\begin{tabular}{|l||c c||c c||}
                \hline
                & \multicolumn{2}{c||}{$\pi(\bfx|\bfy_{\rm proj})$} & \multicolumn{2}{c||}{$\pi_{\rm proj}(\bfx_{\rm proj}|\bfy_{\rm proj})$}\\
			\hline
			  Approach & PCP-Map & CP-Flow & PCP-Map & CP-Flow \\
                Number of Parameters & $\sim$8.9M & $\sim$5.7M & $\sim$2.5M & $\sim$1.4M \\
                Training Mean NLL & $-540.3\pm 4.5$ & $-534.2\pm 9.1$ & $-5.8\pm 0.6$ & $-6.6\pm0.3$\\ 
                Validation Mean NLL & $-519.0\pm 4.1$ & $-506.4\pm 5.1$ & $8.5\pm 0.1$ & $6.1\pm 0.2$ \\ 
			\hline\hline
			Training(s)    & \bf{6652.2$\pm$1030.3} & 47207.4$\pm$7221.0 & \bf{706.9$\pm$61.0} & 1989.4$\pm$10.5\\
			Validation(s)   & \bf{108.5$\pm$17.1} & 232.1$\pm$35.5 & \bf{6.6$\pm$0.6} & 10.3$\pm$0.1\\
            \hline
			Total(s) & \bf{6760.7$\pm$1047.4} & 47439.5$\pm$7256.4 & \bf{713.5$\pm$61.5} & 1999.7$\pm$10.5\\[2pt]
			\hline
		\end{tabular}
        }
	\end{center}
\end{table}

\section{Discussion}
\label{sec:discussion}

In our numerical experiments, the comparison to the SMC-ABC approach for the stochastic Lotka--Volterra problem illustrates common trade-offs when selecting conditional sampling approaches.
Advantages of the ABC approach include its strong theoretical guarantees and well-known guidelines for choosing the involved hyper-parameters (annealing, burn-in, number of samples to skip to reduce correlation, etc.). The disadvantages are that ABC typically requires a large number of likelihood evaluations to produce (approximately) i.i.d.\thinspace samples and low-variance estimators in high-dimensional parameter spaces; the computation is difficult to parallelize in the sequential Monte Carlo setting, and the sampling process is not amortized over the conditioning variable $\bfy^*$, i.e., it needs to be recomputed whenever $\bfy^*$ changes.

Comparisons to the flow-based NPE method~\cite{GATSBI} for the high-dimensional 1D shallow water equations problem (\Cref{sec:shallow}) illustrate the superior accuracy achieved by our approaches. The results indicate that both methods can learn high-dimensional COT maps. Here, the number of effective parameters in the dataset was $n=14$, and the number of effective measurements was $m=3500$. Particularly worth noting is that PCP-Map, on average, converges in around 715 seconds on one GPU on this challenging high-dimensional problem. Moreover, when compared to the amortized CP-Flow approach, PCP-Map achieves significantly faster convergence while providing a working computational scheme to the static COT problem.

Learning posterior distributions using our techniques or similar measure transport approaches is attractive for real-world applications where samples from the joint distributions are available (or can be generated efficiently), but evaluating the prior density or the likelihood model is intractable. 
Common examples where a non-intrusive approach for conditional sampling can be fruitful include inverse problems where the predictive model involves stochastic differential equations (as in \Cref{sec:stochLV}) or legacy code and imaging problems where only prior samples are available. Our conditional OT approaches can also be integrated into LFI frameworks such as BayesFlow~\cite{radev_2023, tejero-cantero2020sbi} to perform the density estimation.
Another interesting avenue of future work is using our conditional generators in hybrid inference approaches such as \cite{orozco_2023}, which demonstrates that physics-based modeling combined with conditional normalizing flows can improve the scalability and sampling efficiency of LFI for large scale problems

One key advantage of our approaches compared to other transport-based methods that lack approximation targets is they allow for the development of theoretical guarantees. Although this paper focuses on computational algorithms and their empirical validation, we see potential in providing statistical complexity analysis and approximation theoretic analysis for approximating COT maps using our approaches in a conditional sampling context. Encouraging recent progress can already be seen in the theoretical analysis of unconditional OT maps~\cite{wang_2022, divol_2024}.

Given the empirical nature of our study, we paid particular attention to the setup and reproducibility of our numerical experiments.
To show the robustness of our approaches to hyperparameters and to provide guidelines for hyperparameter selection in future experiments, we report the results of a simple two-step heuristic that randomly samples hyperparameters and identifies the most promising configurations after a small number of training steps.

Since both approaches perform similarly in our numerical experiments, we want to comment on some distinguishing factors.
One advantage of the PCP-Map approach is its model simplicity, depending only on three hyperparameters (feature width, context width, and network depth), enabled by identifying clearly the conditional Brenier maps as the target. 
We observed consistent performance for most choices of hyperparameter combination. 
This feature is particularly attractive when experimenting with new problems. 
The limitation is that constraints must be imposed to guarantee partial convexity of the potential. On the other hand, the value function (i.e., the velocity field) in COT-Flow can be designed almost arbitrarily. Thus, the latter approach may be beneficial when new data types and their invariances need to be modeled, e.g., permutation invariances or symmetries, that might conflict with the network architecture required by the direct transport map.
Both approaches also differ in terms of their numerical implementation. Training the PCP-Map via backpropagation is relatively straightforward, but sampling requires solving a convex program, which can be more expensive than integrating the ODE defined by the COT-Flow approach, especially when that model is trained well, and the velocity is constant along trajectories. 
Training the COT-Flow model, however, is more involved due to the ODE constraints and influential hyperparameters for the transport cost and HJB penalties.

Beyond the approaches surveyed and discussed, diffusion models~\cite{song2021scorebased} gained much success recently across many applications. In particular, following the procedure in~\cite[Appendix I.4]{song2021scorebased}, the resulting unconditional generator can be used to obtain various conditional distributions as long as there is a tractable and differentiable log-likelihood function; see, e.g., applications to image generation and time series imputation in~\cite{batzolis2021conditional, Tashiro2021CSDICS}. We omit discussions on diffusion models since the notion of measure transport is not immediately obvious in this setting as the reference samples are evolved stochastically.
Besides diffusion models, flow matching has been recently proposed as an alternative training scheme for continuous normalizing flows~\cite{Lipman_2023} with connections to OT. Since flow matching alleviates the need for time integration, investigating extensions to COT merits further investigation~\cite{kerrigan_2024}.

\bibliographystyle{siamplain}
\bibliography{main}
\end{document}

%% file: ex_shared.tex
\usepackage{lipsum}
\usepackage{amsfonts}
\usepackage{graphicx}
\usepackage{epstopdf}
\usepackage{algorithmic}
\ifpdf
  \DeclareGraphicsExtensions{.eps,.pdf,.png,.jpg}
\else
  \DeclareGraphicsExtensions{.eps}
\fi
\usepackage{tikz}
\usetikzlibrary{positioning}
\tikzstyle{label}=[color=white,
fill=darkgray,opacity=.8,
rectangle,rounded corners,
draw,minimum height=0.65cm]
\usepackage{mathtools}

\newsiamremark{remark}{Remark}
\newsiamremark{hypothesis}{Hypothesis}
\crefname{hypothesis}{Hypothesis}{Hypotheses}
\newsiamthm{claim}{Claim}
\newsiamremark{example}{Example}

\usepackage[textsize=tiny]{todonotes}

\definecolor{darkgreen}{rgb}{.15,.55,0}

\headers{Neural Network Approaches for Conditional Optimal Transport}{Z. O. Wang, R. Baptista, Y. Marzouk, L. Ruthotto, D. Verma.}

\title{Efficient Neural Network Approaches for Conditional Optimal Transport with Applications in Bayesian Inference\thanks{Accepted for publication in SIAM Journal on Scientific Computing (SISC).
\funding{LR's and DV's work was supported in part by NSF awards DMS 1751636, DMS 2038118, AFOSR grant FA9550-20-1-0372, and US DOE Office of Advanced Scientific Computing Research Field Work Proposal 20-023231. RB and YM acknowledge support from the US Department of Energy, Office of Advanced Scientific Computing Research, M2dt MMICC center, award number DE-SC0023187. ZW and YM were supported by the Office of Naval Research, SIMDA (Sea Ice Modeling and Data Assimilation) MURI, award number N00014-20-1-2595 (Dr.\ Reza Malek-Madani and Dr.\ Scott Harper). RB gratefully acknowledges support from the Air Force Office of Scientific Research, Machine Learning and Physics-Based Modeling and Simulation MURI, award number FA9550-20-1-0358, and a Department of Defense Vannevar Bush Faculty Fellowship, award N00014-22-1-2790.}}} 
\author{Zheyu Oliver Wang\thanks{Department of Aeronautics and Astronautics, Massachusetts Institute of Technology, Cambridge, MA 
  (\email{olivrw@mit.edu}, \email{ymarz@mit.edu}).}
\and Ricardo Baptista\thanks{Computing + Mathematical Sciences, California Institute of Technology, Pasadena, CA
  (\email{rsb@caltech.edu}).} 
\and Youssef Marzouk\footnotemark[2] \and Lars Ruthotto\thanks{Department of Mathematics, Emory University, Atlanta, GA (\email{deepanshu.verma@emory.edu}, \email{lruthotto@emory.edu}).} \and Deepanshu Verma\footnotemark[4]}

\usepackage{amsopn}
\newcommand{\hf}{{\frac 12}}

\newcommand{\bfA}{{\bf A}}

\newcommand{\bfI}{{\bf I}}

\newcommand{\bfL}{{\bf L}}

\newcommand{\bfV}{{\bf V}}

\newcommand{\bfX}{{\bf X}}

\newcommand{\bfa}{{\bf a}}
\newcommand{\bfb}{{\bf b}}
\newcommand{\bfc}{{\bf c}}

\newcommand{\bfh}{{\bf h}}

\newcommand{\bfs}{{\bf s}}
\newcommand{\bfx}{{\bf x}}
\newcommand{\bfy}{ {\bf y}}
\newcommand{\bfu}{{\bf u}}
\newcommand{\bfq}{{\bf q}}
\newcommand{\bfp}{{\bf p}}

\newcommand{\bfv}{{\bf v}}
\newcommand{\bfw}{{\bf w}}

\newcommand{\bfz}{{\bf z}}

\newcommand{\bfepsilon}{{\boldsymbol \epsilon}}

\newcommand{\Ex}{{\rm \mathbb{E}}}

\newcommand{\Cov}{{\rm Cov}}

\newcommand{\bfSigma}{{\boldsymbol \Sigma}}

\newcommand{\R}{\ensuremath{\mathbb{R}}}

\DeclareMathOperator*{\argmin}{arg\,min}

\newcommand{\gx}{g}

\newcommand{\Ginv}{\tilde{G}}

\newcommand{\Finv}{\tilde{F}}
\newcommand{\ReLU}{\sigma_{\rm ReLU}}